\documentclass[10pt,twocolumn,letterpaper]{article}

\usepackage{cvpr}              

\usepackage[breaklinks,colorlinks,citecolor=cvprblue,linkcolor=abbrcolor,urlcolor=cvprblue,bookmarks=false]{hyperref}
\usepackage[utf8]{inputenc} 
\usepackage[T1]{fontenc}    
\usepackage{url}            
\usepackage{booktabs}       
\usepackage{amsfonts}       
\usepackage{nicefrac}       
\usepackage{microtype}      
\usepackage{xcolor}         
\usepackage{enumitem}      
\usepackage{tcolorbox}

\usepackage{mathtools}
\usepackage{amsmath}
\usepackage{amsthm}
\usepackage{amsfonts}
\usepackage{thmtools}
\usepackage{thm-restate}
\usepackage[capitalize]{cleveref}

\newcommand{\myparagraph}[1]{\smallskip\noindent\textbf{#1.}}

\DeclarePairedDelimiter{\parens}{\lparen}{\rparen}

\newcommand\mD{{\boldsymbol{D}}}

\newcommand\mI{{\boldsymbol{I}}}

\newcommand\vc{{\boldsymbol{c}}}
\newcommand\vd{{\boldsymbol{d}}}

\newcommand\vf{{\boldsymbol{f}}}

\newcommand\vp{{\boldsymbol{p}}}
\newcommand\vq{{\boldsymbol{q}}}
\newcommand\vr{{\boldsymbol{r}}}
\newcommand\vvs{{\boldsymbol{s}}}

\newcommand\vv{{\boldsymbol{v}}}
\newcommand\vw{{\boldsymbol{w}}}

\newcommand\vz{{\boldsymbol{z}}}

\newcommand\bepsilon{\boldsymbol{\epsilon}}

\newcommand\btheta{\boldsymbol{\theta}}

\newcommand\sR{{\mathbb{R}}}

\crefname{lemma}{lemma}{lemmas}
\Crefname{lemma}{Lemma}{Lemmas}
\crefname{thm}{theorem}{theorems}
\Crefname{thm}{Theorem}{Theorems}
\crefname{assumption}{assumption}{assumptions}
\Crefname{assumption}{Assumption}{Assumptions}

\newcommand{\argmin}{\mathop{\rm argmin}}

\usepackage{makecell}
\usepackage{amsmath}
\usepackage{multirow}   
\usepackage{xspace}     
\usepackage{graphicx}
\usepackage{booktabs}
\usepackage{algorithmic}
\usepackage{wrapfig}
\usepackage[lined,boxed,commentsnumbered,ruled,vlined]{algorithm2e}
\usepackage{tabularx}
\usepackage{pifont}
\usepackage{pgfplots}
\pgfplotsset{compat=1.18}
\usepackage{inconsolata}
\usepackage{times} 

\definecolor{promptlightblue}{HTML}{F1FBFE}
\definecolor{lighterblue}{HTML}{DBE9FC}
\definecolor{darkblue}{HTML}{6C8EBF}

\definecolor{lightgray}{HTML}{D3D3D3}
\definecolor{darkgray}{HTML}{666666}

\definecolor{lightorange}{HTML}{FFE7CC}

\definecolor{lightpurple}{HTML}{E1D6E8}
\definecolor{darkpurple}{HTML}{9674A6}

\definecolor{cvprblue}{rgb}{0.21,0.49,0.74}
\definecolor{abbrcolor}{HTML}{990000}
\definecolor{linkcolor}{rgb}{0.79,0.51,0.26}
\definecolor{jinqilightgray}{HTML}{F5F5F5}
\definecolor{jinqilightorange}{HTML}{FFE7CC}
\definecolor{jinqilightblue}{HTML}{DBE9FC}
\definecolor{jinqilightpurple}{HTML}{E1D6E8}

\definecolor{jinqidarkgray}{HTML}{666666}
\definecolor{jinqidarkorange}{HTML}{D79C01}
\definecolor{jinqidarkblue}{HTML}{6C8EBF}
\definecolor{jinqidarkpurple}{HTML}{9674A6}
\definecolor{jinqidarkgreen}{HTML}{70AD47}
\definecolor{jinqidarkred}{HTML}{990000}

\definecolor{pennred}{HTML}{990000}
\definecolor{pacegreen}{HTML}{00B050}
\definecolor{paceorange}{HTML}{D79C00}

\newcommand{\ourframework}{CoLan\xspace}
\newcommand{\ourdataset}{CoLan-150K\xspace}
\newcommand{\origtt}[1]{{\fontfamily{cmtt}\selectfont #1}}

\SetKwComment{Comment}{$\triangleright$\ }{}

\title{Concept Lancet: Image Editing with Compositional Representation Transplant}

\author{Jinqi Luo, Tianjiao Ding, Kwan Ho Ryan Chan, Hancheng Min, Chris Callison-Burch, René Vidal \\
University of Pennsylvania\\
\origtt{\small jinqiluo@upenn.edu}
}

\begin{document}
\maketitle
\begin{abstract} 
Diffusion models are widely used for image editing tasks. Existing editing methods often design a representation manipulation procedure by curating an edit direction in the text embedding or score space. However, such a procedure faces a key challenge: overestimating the edit strength harms visual consistency while underestimating it fails the editing task. Notably, each source image may require a different editing strength, and it is costly to search for an appropriate strength via trial-and-error. To address this challenge, we propose \textcolor{abbrcolor}{\textbf{Co}}ncept \textcolor{abbrcolor}{\textbf{Lan}}cet (\ourframework), a zero-shot plug-and-play framework for principled representation manipulation in diffusion-based image editing. At inference time, we decompose the source input in the latent (text embedding or diffusion score) space as a sparse linear combination of the representations of the collected visual concepts. This allows us to accurately estimate the presence of concepts in each image, which informs the edit. Based on the editing task (replace/add/remove), we perform a customized concept transplant process to impose the corresponding editing direction. To sufficiently model the concept space, we curate a conceptual representation dataset, \ourdataset, which contains diverse descriptions and scenarios of visual terms and phrases for the latent dictionary. Experiments on multiple diffusion-based image editing baselines show that methods equipped with \ourframework achieve state-of-the-art performance in editing effectiveness and consistency preservation. More project information is available at \href{https://peterljq.github.io/project/colan}{https://peterljq.github.io/project/colan}.
\end{abstract}
    
\vspace{-3mm}
\section{Introduction}
\label{sec:intro}

\begin{figure}[t]
\centering
  \includegraphics[width=\linewidth]{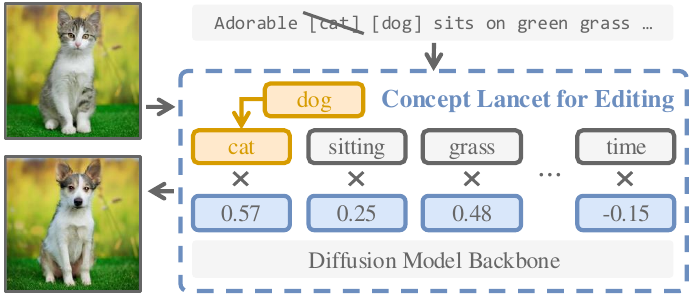}
\vspace{-5mm}
\caption{Given a source image and the editing task, our proposed \ourframework generates a concept dictionary and performs sparse decomposition in the latent space to precisely transplant the target concept.
}
\label{fig:teaser_figure}
\vspace{-5mm}
\end{figure}

 How can we edit a given image following a specified conceptual guidance, say \texttt{Cat$\rightarrow$Dog} or \texttt{Sketch$\rightarrow$Painting}? This problem was briefly discussed in the early work of \emph{Image Analogies} \cite{hertzmann2001image} in the 2000s, widely studied in the era of generative adversarial networks \cite{2019stylegan,CycleGAN2017}, and recently revolutionized by diffusion models \cite{ddpm,sohldickstein2015deep}. Not only do diffusion models shine in producing realistic images \cite{saharia2022imagen,Peebles2022DiT}, but they also allow for conditioning on multimodal guidance such as text prompts \cite{rombach2021highresolution,khachatryan2023text2videozero} and regional masks \cite{Avrahami_2022_CVPR_blended_latent_diffusion,zhang2023adding,huang2023pfbdiff}, making it handy for image editing.

In this paper, we consider the task of utilizing diffusion models for image editing that impose desired concepts based on user prompts. Specifically, given a source image and its source caption, our task is to modify the content, appearance, or pattern of the image based on a given target prompt. To accomplish it with diffusion models, a basic idea is to first use the score predictor to perform the noising process (e.g., DDIM Inversion \cite{ddim}) and then follow the denoising process conditioned on the target concept. Such conditioning in the diffusion-based backbones typically happens in a structured latent space (score or text embedding space; to be detailed in \S\ref{sec:basics}), where one moves a source latent vector towards an edit direction (i.e., the shift from the source concept to the target concept). A fruitful series of works have contributed to enhancing the inversion process \cite{song2023consistency,ju2023directinversion,li2023stylediffusion,mokady2022nulltext,huberman2024editfriendly}, innovating attention controls \cite{hertz2023prompttoprompt,parmar2023pix2pixzero,xu2023inversion,cao2023masactrl}, and tuning the backbone with conditional instructions \cite{ouyang2022RLHF,Brooks_2023,geng2023instructdiffusion,zhuang2023task}.

\begin{figure}[t]
\centering
  \includegraphics[width=\linewidth]{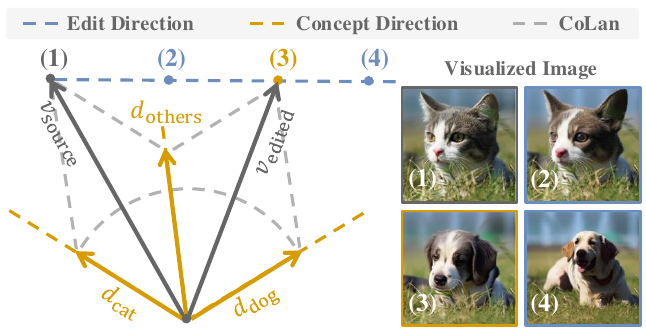}
\caption{Representation manipulation in diffusion models involves adding an \textit{accurate} magnitude of edit direction (e.g., Image \textcolor{jinqidarkorange}{(3)} by \ourframework) to the latent source representation. Figure~\ref{fig:visual_comparison_p2pzero} and Figure~\ref{fig:visual_comparison_infedit} show more examples.
}
\label{fig:vector_space}
\vspace{-3mm}
\end{figure}

Despite the remarkable progress, an often overlooked issue is the \textit{magnitude} of the concept representation to be imposed, that is, determining how far to move along the edit direction. Prior works \cite{parmar2023pix2pixzero,xu2023inversion} typically assume a certain amount of editing without adjusting the magnitude based on the contexts or existing concepts of the given image. Such a heuristic choice can be problematic as we show in \Cref{fig:vector_space}: adding too much of the edit may overwhelm the image with the target concept (Image (4)), while too little may fail to fully remove the source concept (Image (2)). 
Hence, since the presence of the source concept varies in every image, proper estimation of the edit magnitude is necessary for accurate editing. 

To address the aforementioned issue, we draw inspiration from low-dimensional \cite{liang2024how,2022diffusionclip} and compositional \cite{NEURIPS2023_scorealgebra,Li_2024_latent_direction} structures in latent representations of diffusion models. In this paper, we propose Concept Lancet (\ourframework), a zero-shot plug-and-play framework to interpret and manipulate sparse representations of concepts for diffusion-based image editing. Our intuition is to sufficiently model the latent spaces to analyze how much each concept is present in the source representation, which allows for accurate transplant from the source concept to the target one in the proper magnitude. We state our contributions and workflows as follows: 
\begin{itemize}[wide,itemindent=5pt]
    \item (\S \ref{sec:method-gendict}) To allow for such an analysis, one needs a dictionary of directions representing diverse concepts. Existing dictionaries are limited since they either contain concepts without clear visual meanings (e.g., ``hardship''), lack phrases (e.g., ``made of wood''), or have only a limited number of concepts. Thus, we collect a conceptual representation dataset, \ourdataset, of diverse descriptions for visual concepts and compute a dictionary of concept vectors to the latent (text embedding or score) space.
    \item (\S \ref{sec:method-transplant}) At inference time, we propose to decompose the source latent vector as a linear combination of the collected visual concepts to inform the edit. To mitigate the optimization inefficiency with an overcomplete dictionary, 
    we instruct a vision-language model (VLM) to parse image-prompt tuples into a representative list of visual concepts as dictionary atoms. For common editing tasks of replacing, we switch the source concept vector in the decomposition with our target concept vector and synthesize the edited image with the backbone. 
The task of adding or removing concepts can be recasted as special cases of concept replacing, detailed in the method section.
    \item (\S \ref{sec:exp}) We conduct quantitative comparisons on multiple diffusion-based image editing baselines and qualitative evaluations on the visual synthesis. Methods equipped with \ourframework achieve state-of-the-art performance in editing effectiveness and consistency preservation. Notably, the plug-and-play design of our method provides flexibility in the choice of backbones and latent spaces.
\end{itemize}

\vspace{-1mm}
\section{Preliminaries in Diffusion-Based Editing}
\vspace{-1mm}
\label{sec:basics}
This section briefly discusses diffusion-based image editing and how it involves a representation manipulation process in either the text embedding or the score space.

\myparagraph{DDIM Inversion} Diffusion model samples a new image $\vz_0$ by gradually denoising a standard Gaussian $\vz_T$ through the following reverse-time conditional denoising process:
\begin{align}
    \vz_{t-1}&=\sqrt{\alpha_{t-1}} \cdot \parens*{\frac{\vz_t-\sqrt{1-\alpha_t}\bepsilon_{\btheta}(\vz_t,t,\vc)}{\sqrt{\alpha_t}}}\nonumber\\
    &\quad + \sqrt{1-\alpha_{t-1}-\sigma_t^2} \cdot \bepsilon_{\btheta}(\vz_t,t,\vc)\nonumber\\
    &\quad + \sigma_t \bepsilon_t,\ \text{ with } \bepsilon_t\sim\mathcal{N}(0,\mI)\,,\label{eq_denoise}
\end{align}
where $\vz_t$ is the denoised image at time $t$ $(t=0,\dots,T)$, $\vc$ is the text embedding of the caption of the image to be sampled,
and $\bepsilon_{\btheta}(\vz_t,t,\vc)$ models the \emph{score function}~\cite{song2021scorebased} for noise prediction. 
 With the choice of $\sigma_t=0$, the denoising process in \eqref{eq_denoise} enables DDIM inversion~\cite{ddim}. 
Specifically, one replaces the forward process with the iterative calling of $\bepsilon_{\btheta}(\cdot,t,\vc)$ to predict the forward noise for the source $\vz_0$. The forward process often stores anchor information (e.g., attention map) \cite{hertz2023prompttoprompt,parmar2023pix2pixzero,xu2023inversion,cao2023masactrl}.
Then the reverse process samples the \emph{target image} $\vz_0'$ with visual features corresponding to altered concepts. 
The key ideas are first to utilize the anchor information such that the denoising process can faithfully reconstruct $\vz_0$ when conditioned on $\vc$ (often called sampling in the reconstruction/source branch), and then to further implant \eqref{eq_denoise} with a representation manipulation procedure. 

For simplicity, we describe the basic DDIM paradigm here. We further elaborate variants of inversion methods for exact recovery of $\vz_0$ \cite{ju2023directinversion, xu2023inversion} and different diffusion backbones (e.g., Stable Diffusion \cite{rombach2021highresolution,podell2023sdxl}, Consistency Model \cite{song2023consistency,luo2023latentconsistency}) in Appendix \S\ref{sec:appendix_prior_art} and \S\ref{sec:appendix_implementation_details}.

\myparagraph{Steering Reverse Process} Prior works have different choices of representation manipulation in the space of text embeddings \cite{radford2021clip} or diffusion scores \cite{luo2022understanding} to vary visual synthesis. A general paradigm is to impose an editing direction ($\Delta\bepsilon$ or $\Delta\vc$) to the latent representation (the score function prediction $\bepsilon_{\btheta}(\cdot;t,\vc)$ or the text embedding $\vc$, respectively) in the editing backbone. Sampling with manipulated concepts is often called sampling in \emph{edit/target branch}. 

\begin{figure*}[t]
    \centering
    \includegraphics[width=\linewidth]{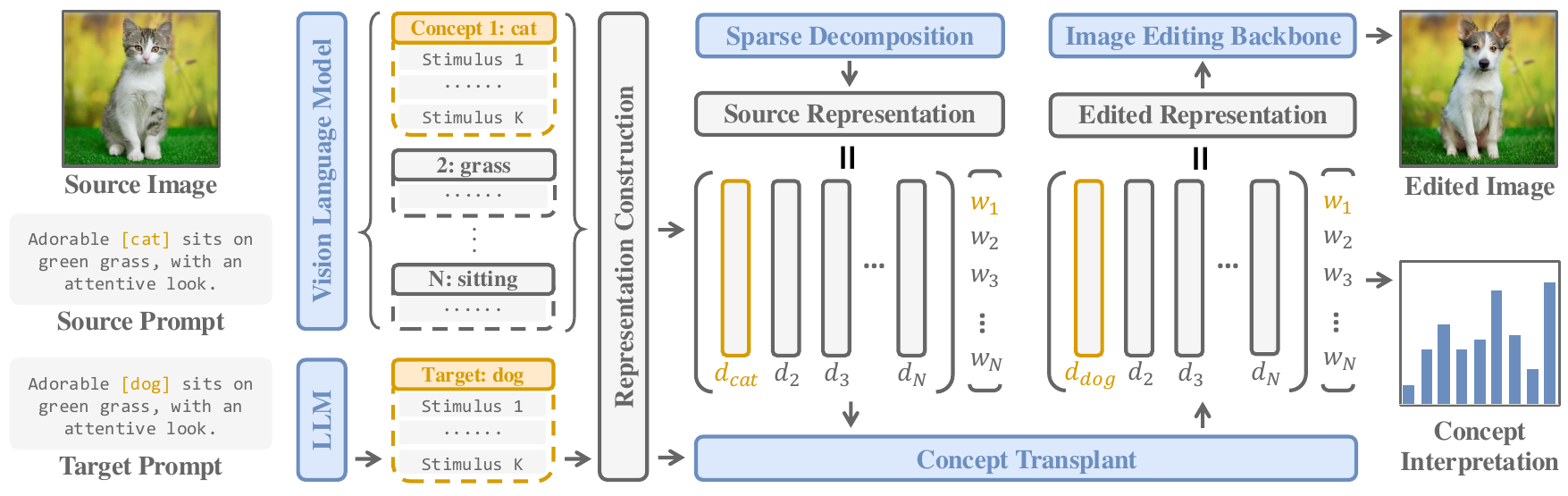}
    \vspace{-6mm}
    \caption{The \ourframework framework. Starting with a source image and prompt, a vision-language model extracts visual concepts (e.g., cat, grass, sitting) to construct a concept dictionary. The source representation is then decomposed along this dictionary, and the target concept (dog) is transplanted to replace the corresponding atom to achieve precise edits. Finally, the image editing backbone generates an edited image where the desired target concept is incorporated without disrupting other visual elements. }
    \label{fig:method}
    \vspace{-3mm}
\end{figure*}
\begin{itemize}[wide,itemindent=5pt]
    \item \textit{(Text Embedding Space)} For instance, the work of \cite{parmar2023pix2pixzero} 
edits the concept by manipulating the $\vc$ in Equation \ref{eq_denoise} as 
\begin{equation*}
    \vc_{\text{edit}} = \vc_{\text{source}} + w \left(\vc_{\text{A}}-\vc_{\text{B}}\right),
\end{equation*}
where $\Delta \vc = \vc_{\text{A}}-\vc_{\text{B}}$ is a generalized direction denoting the shift from concept B towards concept A. In fact, this formulation of Vector Addition (VecAdd) can be traced back to word embeddings in language models: the work of \cite{mikolov_linguistic_2013} observes that $\vc_{\text{queen}}\approx\vc_{\text{king}}+w \cdot \left(\vc_{\text{woman}}-\vc_{\text{man}}\right)$, where $\vc_{\text{king}}$ can be viewed as a source latent vector and $\vc_{\text{woman}}-\vc_{\text{man}}$ as an editing direction. Recent years have also witnessed such an idea being applied to steering activations of large language models (LLMs) \cite{alex2023actadd,zou2023representation,luo2024pace} and VLMs \cite{tewel2022zerocap, couairon2022embedding, trager2023linear,liu2024reducing}.

\item \textit{(Score Space)} Such an idea has also been applied in the score space by adding a score $\bepsilon$ with a $\Delta \bepsilon$. For instance, the work of \cite{ho2022classifier,Gandikota_2023, Brack_2024_CVPR} considered the recipe \begin{equation}
\bepsilon_{\text{edit}} = \bepsilon_{\btheta}(\cdot;t,\vc_{\text{source}}) + w (\bepsilon_{\btheta}(\cdot;t,\vc_{\text{target}})-\bepsilon_{\btheta}(\cdot;t,\vc_{\text{source}})),\nonumber
\end{equation}
where $\bepsilon_{\btheta}(\cdot;t,\vc_{\text{source}})$ can be treated as the source latent vector $\bepsilon$, $\bepsilon_{\btheta}(\cdot;t,\vc_{\text{target}})-\bepsilon_{\btheta}(\cdot;t,\vc_{\text{source}})$ as the edit direction $\Delta \bepsilon$, and $w$ controls the amount of edit. One can also implement $\vc_{\text{source}} = \emptyset$ to have the unconditional score. More generally, there is a line of works \cite{xu2023inversion,Gandikota_2023,ho2022classifier,wu2022unifying,NEURIPS2023_scorealgebra,chen2024exploring} involving the above formulation to steer or edit the synthesis.

\end{itemize}

\section{Our Method: Concept Lancet}
With the above background we are ready to propose our method for accurate representation manipulation in diffusion-based image editing. The high-level idea is that, instead of arbitrarily setting the amount of edit, we will estimate what and how much concepts are present in the source image to inform the edit. This is done via collecting a dictionary of concept vectors in the latent space and decomposing the source latent vector into a linear combination of the dictionary atoms to allow the concept transplant procedures, which we shall discuss in \S \ref{sec:method-gendict} and \S \ref{sec:method-transplant} respectively.

\subsection{Concept Dictionary Synthesis} \label{sec:method-gendict}
Here the main goal is to collect a diverse set of concepts (and the corresponding concept vectors in the latent space) that are both visually meaningful and relevant for image editing, such that the decomposition of a source latent vector captures important visual elements and allows potential modifications for effective editing. This naturally boils down to two steps: curating visual concepts for stimulus synthesis and extracting a concept vector from the stimuli. We describe our approach below and compare it with the alternatives in the literature. 

\begin{figure*}[t]
    \centering
    \includegraphics[width=\linewidth]{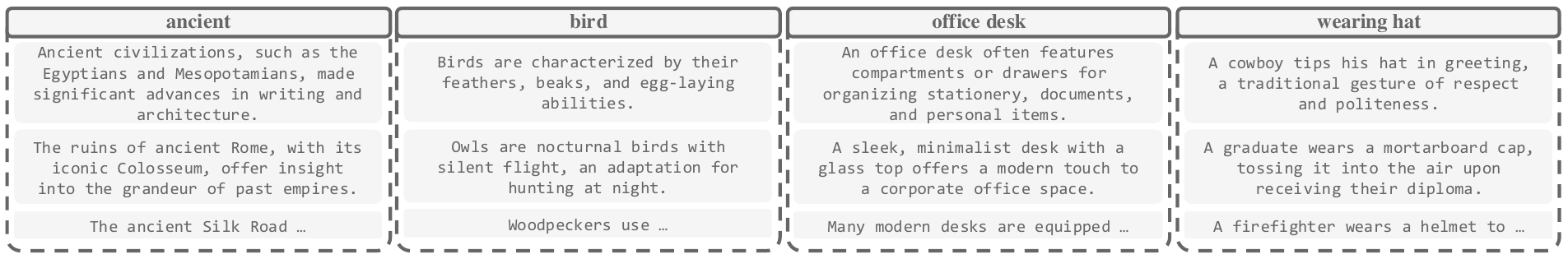}
    \vspace{-6mm}
    \caption{Samples of the concept stimuli from \ourdataset. Additional samples are attached in the Appendix \S\ref{sec:appendix_additional_results}.}
    \label{fig:concept_stimulus}
    \vspace{-4mm}
\end{figure*}

\myparagraph{Curating Visual Concepts}  
Constructing domain-specific concepts is widely adopted for evaluating and controlling generative foundation models \cite{li2024logicity,li2024embodied,liu2022compositional,wu2024conceptmix,wu2025unlearning,lee2023holistic_heim}. To model the rich semantics of a given concept, an emerging line of work collects textual concept stimuli (i.e., a set of examples, descriptions, and scenarios) for downstream LLM or diffusion editing tasks \cite{parmar2023pix2pixzero,zou2023representation,alex2023actadd,luo2024pace}.  
There are three issues when applying these concepts in editing images:
\begin{itemize}[wide,itemindent=5pt]
    \item Many concepts for editing LLMs \cite{zou2023representation,luo2024pace}, such as ``honesty'' or ``hardship,'' are not catered to image editing in diffusion models. Existing concept stimuli are typically in a specialized format for LLM activation reading (e.g., begin with second-person pronouns).
    \item Such concepts primarily focus on single-word descriptor (e.g. ``love'', ``friendship''), rather than multi-word phrases (e.g., ``wearing sunglasses'' or ``made of wood'') that are helpful to model visual space. 
    \item Existing collection of concepts for image editing has a limited number of concept vectors open-sourced (e.g., less than 20 in \cite{parmar2023pix2pixzero} and less than 50 in \cite{2021StyleCLIP,luo2023zeroshot}).
\end{itemize}
To address these issues, we curate a comprehensive set of visual concepts relevant to image editing tasks. 
Specifically, for each editing task that consists of a source image, a source prompt, and an editing prompt, we employ a VLM \cite{gpt4v_system_card} to parse the image prompts tuple and generate a list of relevant visual concepts. This step ensures that our concepts are both visually grounded and editing-relevant. 

We then instruct an LLM \cite{openai2023gpt4} with in-context demonstrations of stimulus synthesis to generate diverse stimuli for each concept to capture various contexts in which the concept shows up. The instructions are shown in Appendix \S\ref{sec:appendix_prompting_template}. After collecting concepts across all editing tasks, we obtain $5,078$ concepts and a total of $152,971$ concept stimuli, which we call \ourframework-150K. Figure~\ref{fig:concept_stimulus} shows samples of the concept stimuli in our dataset.
Compared to existing collections of conceptual representation for diffusion-based editing, \ourdataset represents a significant scaling up and provides richer and more diverse representations for each concept. 
By sampling various observations of a concept, the large set of stimuli helps accurately estimate a representation that is robust to changes in contexts.

\myparagraph{Concept Vector Extraction} 
Given the stimuli for each concept, we now need to extract a representative direction in the latent space. Let $x$ be a concept (e.g., ``wearing sunglasses'') and ${\vvs_1^x, \dots, \vvs_K^x}$ be its corresponding stimuli. We first map each stimulus into the latent space using the text encoder $E$ of the diffusion model\footnote{For simplicity, we describe the concept extraction in the text embedding space; see the Appendix \S\ref{sec:appendix_implementation_details} for the case of the score space.}. To read a robust concept vector from the collection of embeddings of stimuli, we draw inspiration from prior arts on Representation Reading ($\operatorname{RepRead}$) and propose two options: Arithmetic Average ($\operatorname{Avg}$) \cite{zou2023representation,luo2024pace,parmar2023pix2pixzero,subramani_extracting_2022} or Principal Component Analysis ($\operatorname{PCA}$) \cite{liu2024reducing,zou2023representation,luo2024pace} on the set of embedding vectors. $\operatorname{Avg}$ directly returns the mean of all stimulus embeddings and $\operatorname{PCA}$ returns the first principal component of embeddings as the concept vector: 
\begin{equation}
\vd_x = \operatorname{RepRead}({E(\vvs_1^x), \dots, E(\vvs_K^x)}).
\end{equation}
For each given source image, a specific collection of concept vectors $\{\vd_{x_i}\}_{i=1}^{N}$ will form the concept dictionary, which will be further used for decomposition analysis during inference (\S \ref{sec:method-transplant}). Figure~\ref{fig:concept_stimulus} shows samples of concepts and their associated stimuli. We use $\operatorname{Avg}$ for the representation reading stage in the experiments since it is more computationally efficient.   

\subsection{Concept Transplant via Sparse Decomposition} \label{sec:method-transplant}
Now that we have obtained a concept dictionary, we are ready to describe how we decompose the latent code of the image along the dictionary and transplant the concept. 

\myparagraph{Selecting Task-Specific Concepts} While our concept dictionary provides a comprehensive collection of visual concepts, not all concepts are relevant to a specific editing task. To avoid spurious decompositions and make the method efficient, the VLM parses the source image-prompt pair and identifies pertinent task-relevant concepts, as we have done in \S \ref{sec:method-gendict}. The corresponding concept vectors are then assembled into a dictionary matrix $\mD \in \sR^{d\times N}$, where $d$ is the dimension of the latent space and $N$ is the number of concepts in the dictionary. More details of constructing the dictionary in a specific latent space (e.g., CLIP text embedding space) are shown in Appendix \S\ref{sec:appendix_implementation_details}.

\myparagraph{Concept Analysis} Given a source latent vector $\vv$ (either from the text encoder or score function), we decompose it along the directions in $\mD$ through sparse coding. That is, we solve the following optimization problem:
\begin{equation}
\vw^* = \argmin_{\vw} \left\|\vv - \mD\vw\right\|_2^2 + \lambda\left\|\vw\right\|_1
\end{equation}
where solutions of concept coefficients $\vw\in\sR^n$ and $\lambda > 0$ is a regularization parameter that controls the sparsity of the solution. In practice, we realize the sparse solver with Elastic Net \cite{you_oracle_2016}. Such a decomposition yields
\begin{equation}
\vv = \mD\vw^* + \vr
\end{equation}
where $\vw^*$ contains the solved coefficients of each concept vector for composition and $\vr$ is the residual not explained by the concepts in $\mD$. 

\begin{table*}[t]
\centering
\footnotesize
\caption{\label{tab:concept_lancet_performance} Evaluation of different baselines using Concept Lancet or Vector Addition. The best performance of each category is in \textbf{bold} and the second best is \underline{underlined}. For each metric under Consistency Preservation, the number on the left is evaluated on the whole image, and the number on the right is evaluated on the background (outside the edit mask).}
\vspace{-2mm}
\begin{tabularx}{\textwidth}{@{}cc>{\centering\arraybackslash}p{1.75cm}%
*{8}{>{\centering\arraybackslash}p{0.7cm}}%
>{\centering\arraybackslash}X%
>{\centering\arraybackslash}X@{}}
 \toprule
 \multirow{3}{*}{\makecell{Representation\\Manipulation}} & \multirow{3}{*}{\makecell{Inversion}} & \multirow{3}{*}{Backbone} & \multicolumn{8}{c}{Consistency Preservation}     & \multicolumn{2}{c}{Edit Effectiveness (\%, $\uparrow$)}  \\
 \cmidrule(lr){4-11} \cmidrule(lr){12-13}
  & & & \multicolumn{2}{c}{\multirow{2}{*}{\makecell{StruDist\\($\times10^{-3}, \downarrow$)}}}   & \multicolumn{2}{c}{\multirow{2}{*}{\makecell{PSNR ($\uparrow$)}}} & \multicolumn{2}{c}{\multirow{2}{*}{\makecell{LPIPS\\($\times10^{-3}, \downarrow$)}}}   &  \multicolumn{2}{c}{\multirow{2}{*}{\makecell{SSIM\\(\%, $\uparrow$)}}} & \multirow{2}{*}{\makecell{Target\\Image}} & Target Concept  \\
 \midrule
  \multirow{1}{*}{\rotatebox[origin=c]{0}{N.A.}} 
                              & \multirow{1}{*}{\rotatebox[origin=c]{0}{DDIM}} & P2P \cite{hertz2023prompttoprompt}               & 69.01 & 39.09 & 15.04 & 17.19 & 340.3  & 221.3  &  56.56 & 70.36 & 24.35 & 21.10  \\
   \multirow{1}{*}{\rotatebox[origin=c]{0}{N.A.}} & \multirow{1}{*}{\rotatebox[origin=c]{0}{DI}}  & P2P \cite{hertz2023prompttoprompt}                & \textbf{11.02} & \textbf{5.963} & 22.71 & 27.24 &  \underline{114.1}   &  \underline{54.68} &  75.08  & 84.57 &  24.82  &  22.07  \\
   \multirow{1}{*}{\rotatebox[origin=c]{0}{N.A.}}  &  \multirow{1}{*}{\rotatebox[origin=c]{0}{DI}} & MasaCtrl \cite{cao2023masactrl}           & 23.34 &  10.40  & 19.12 & 22.78 &  160.8  & 87.38 &  71.12  & 81.36 &   24.42 & 21.37 \\
 \midrule
  \multirow{1}{*}{\rotatebox[origin=c]{0}{VecAdd}} 
                              &  \multirow{1}{*}{\rotatebox[origin=c]{0}{DI}} & P2P-Zero \cite{parmar2023pix2pixzero}           & 53.04 & 25.54  & 17.65 &  21.59 & 273.8  & 142.4 & 61.78  & 76.60 &  23.16 & 20.81 \\
\multirow{1}{*}{\rotatebox[origin=c]{0}{\makecell{\ourframework}}}  
                              & \multirow{1}{*}{\rotatebox[origin=c]{0}{DI}}  & P2P-Zero \cite{parmar2023pix2pixzero}            & 15.91 &  6.606 & \underline{23.08} & 26.08 &  120.3 &  68.43 & \textbf{75.82}  & 83.55 &   23.84  & 21.13 \\
 \midrule
\multirow{1}{*}{\rotatebox[origin=c]{0}{VecAdd}}                            
                              &  \multirow{1}{*}{\rotatebox[origin=c]{0}{VI}} & InfEdit \cite{xu2023inversion}           & 27.18 & 17.24  & 21.83  & 27.99  & 136.6  & 56.65  & 71.70  & 84.64 &  24.80 & 22.04 \\   
\multirow{1}{*}{\rotatebox[origin=c]{0}{\makecell{\ourframework}}} 
                              & \multirow{1}{*}{\rotatebox[origin=c]{0}{VI}}  & InfEdit (E) \cite{xu2023inversion}       & 16.21 &  8.025  & 22.13 & \underline{28.04} & 125.9 & 55.05 & 74.96  & \underline{84.72}&  \underline{24.90} & \underline{22.12} \\
\multirow{1}{*}{\rotatebox[origin=c]{0}{\makecell{\ourframework}}} 
                              & \multirow{1}{*}{\rotatebox[origin=c]{0}{VI}}  & InfEdit (S) \cite{xu2023inversion}       &\underline{13.97} &  \underline{6.199} & \textbf{23.42} & \textbf{28.46} &  \textbf{110.3}  & \textbf{53.04} & \underline{75.51}  & \textbf{85.12}  & \textbf{24.94} & \textbf{22.45} \\
 \bottomrule
\end{tabularx}
\vspace{-2mm}
\end{table*}


\myparagraph{Concept Transplant} To perform the representation manipulation, we construct a modified dictionary $\mD'$ by replacing the column of the source concept vector with that of the target concept. The edited latent representation is then obtained as $\vv' = \mD'\vw^* + \vr$. This transplant scheme preserves the compositional coefficients estimated from the source representation while substituting the relevant concept vector. It imposes the desired concept while maintaining the overall structure of the remaining concepts in the source image.

We note that this concept replacing scheme generalizes to concept insertion and removal. Indeed, concept removal can be viewed as setting the target concept as the null concept; we extract a direction for the null concept using the same procedure as described in \S \ref{sec:method-gendict} with stimuli as empty sentences. On the other hand, the case of concept insertion is more subtle since there is no explicit source concept to replace. Hence we instruct the VLM to comprehend the source image and the target prompt
to suggest an appropriate source concept as the counterpart of the target concept. For example, if the task is to add concept \texttt{[rusty]} to an image of a normal bike, the VLM will identify the concept \texttt{[normal]} for the concept dictionary and the following replacement. 

\section{Experimental Results} \label{sec:exp}

We provide quantitative evaluations for baselines in \S\ref{sec:quantitative} and qualitative observations in \S\ref{sec:qualitative}. Finally, we provide visual analysis of the concept vectors from \ourdataset in \S\ref{sec:rep-analysis}.

\begin{figure*}[t]
    \centering
    \includegraphics[width=\linewidth]{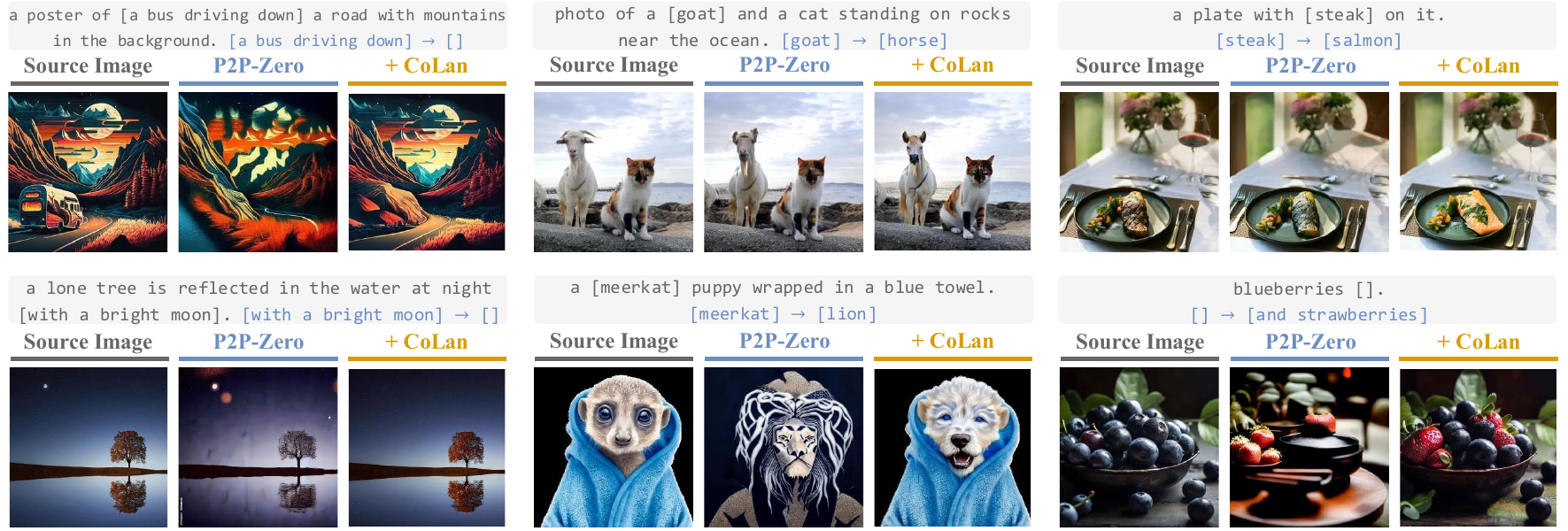}
    \vspace{-6mm}
    \caption{Visual comparisons of \ourframework in the text embedding space of P2P-Zero.  \texttt{\textcolor{jinqidarkgray}{Texts in gray}} are the original captions of the source images from PIE-Bench, and \texttt{\textcolor{jinqidarkblue}{texts in blue}} are the corresponding edit task (replace, add, remove). \texttt{[x]} represents the concepts of interest, and \texttt{[]} represents the null concept.}
    \label{fig:visual_comparison_p2pzero}
    \vspace{-1mm}
\end{figure*}

\begin{figure*}[t]
    \centering
    \includegraphics[width=\linewidth]{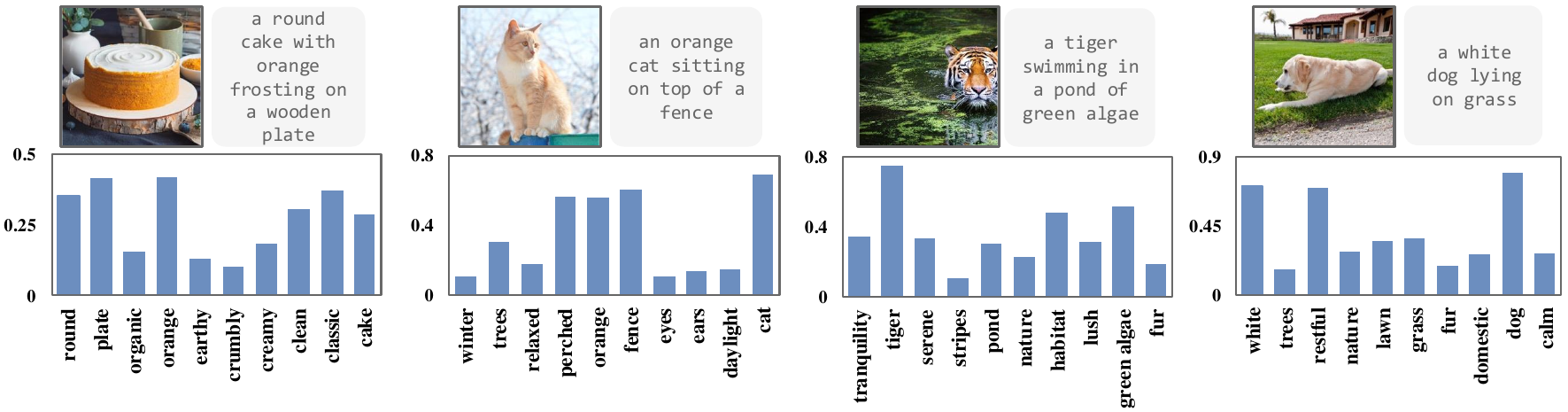}
    \vspace{-6mm}
    \caption{The histograms of solved magnitudes of the concept atoms in \ourframework decomposition (text embedding space). As there are tens of concepts in a single dictionary, the histogram includes the concepts whose \ourframework coefficients have the top 10 largest magnitudes.}
    \label{fig:coefficient_clip}
    \vspace{-1mm}
\end{figure*}

\subsection{Quantitative Evaluation}\label{sec:quantitative}

We perform a standardized quantitative evaluation of \ourframework against current methods with PIE-Bench \cite{ju2023directinversion}. Its editing tasks are based on a broad collection of image sources (e.g., TEdBench \cite{kawar2023imagic}, TI2I benchmark \cite{Tumanyan_2023_CVPR}) with diverse scene types and editing categories.

\myparagraph{Baselines} We compare editing backbones that fall into two categories based on their concept transfer approach: (1) mechanistic swap of attention maps including P2P \cite{hertz2023prompttoprompt} and MasaCtrl \cite{cao2023masactrl}, and (2) representation manipulation that enables us to plug \ourframework in the diffusion score space (S) of InfEdit~\cite{xu2023inversion} and the text embedding space (E) of both InfEdit and P2P-Zero \cite{parmar2023pix2pixzero}.
We cover multiple inversion approaches such as DDIM \cite{ddim}, Direct Inversion (DI) \cite{ju2023directinversion}, and Virtual Inversion (VI) \cite{xu2023inversion}. Further implementation details can be found in Appendix \S\ref{sec:appendix_implementation_details}.

\myparagraph{Metrics} The two main criteria are \textit{Consistency Preservation} and \textit{Edit Effectiveness}. Consistency Preservation is a set of metrics aimed at evaluating the amount of semantic information preserved during image editing. We report the Structure Distance (StruDist) \cite{Tumanyan_2022_CVPR_structure_distance}, PSNR \cite{PSNR}, LPIPS \cite{lpips_cvpr}, and SSIM~\cite{ssim_tip}. On the other hand, Edit Effectiveness measures the correctness of the edited part, and it is evaluated by two metrics: Target Image metric computes the CLIP similarity \cite{wu2021godiva,radford2021clip} between the edited text and the edited image, whereas Target Concept metric computes the CLIP similarity between the edited text and the edit-masked region of the target image.

\myparagraph{Results} Table~\ref{tab:concept_lancet_performance} reports our results.  All backbones equipped with \ourframework have improved Edit Effectiveness, which indicates that \ourframework accurately edits images towards the desired target concept. Moreover, we observe that backbones with \ourframework achieve better consistency preservation across the board. For instance, on the P2P-Zero backbone, \ourframework is able to achieve a lower StruDist and LPIPS by nearly 50\% and a higher PSNR and SSIM by about 10\%. While DI with P2P achieves the best StruDist, \ourframework ranks a very close second for StruDist and overall achieves better performance on all rest of the consistency metrics. We argue that StruDist computes an average difference between the DINO-V2 feature maps of the two images. Hence this single metric is largely dependent on a specific transformer, and checking holistically four metrics is a fairer way for consistency evaluation. Notably, InfEdit with \ourframework in the score space has the most outstanding performance across multiple metrics.  

Additionally, Table~\ref{tab:time_efficiency} shows the average time of sparse decomposition of \ourframework using the CLIP space of InfEdit and P2P-Zero backbones. We observe that, since VLM helps make the dictionary concise, the decomposition only occupies a small proportion of the total editing time. This demonstrates that \ourframework is efficient and inexpensive relative to the overall computation cost of inference in diffusion models. Furthermore, Table~\ref{tab:dictionary_size} compares the editing performance of \ourframework given different dictionary sizes. As expected, we observe that a larger \ourframework dictionary is better at capturing the presence of existing concepts in the source image, leading to stronger editing performance. Overall, our quantitative experiments demonstrate that the concept transplant process of \ourframework benefits from proper accurate and sparse concept representations that exist in the CLIP space and the diffusion score space for better image editing performance.

\vspace{-1mm}
\subsection{Qualitative Observation}\label{sec:qualitative}
\vspace{-1mm}
This section provides qualitative results of edited images. We compare the visual quality between images edited with a given backbone and that complemented with \ourframework.
\subsubsection{Visual Comparison}
Each target image can be segmented into two parts: i) the region of interest, which corresponds to the source concept and should be edited to express the target concept; and ii) the background, whose contents should be intact through the editing process. Here, we qualitatively analyze these two aspects when using \ourframework for image editing. 

\begin{figure*}[t]
    \centering
    \includegraphics[width=\linewidth]{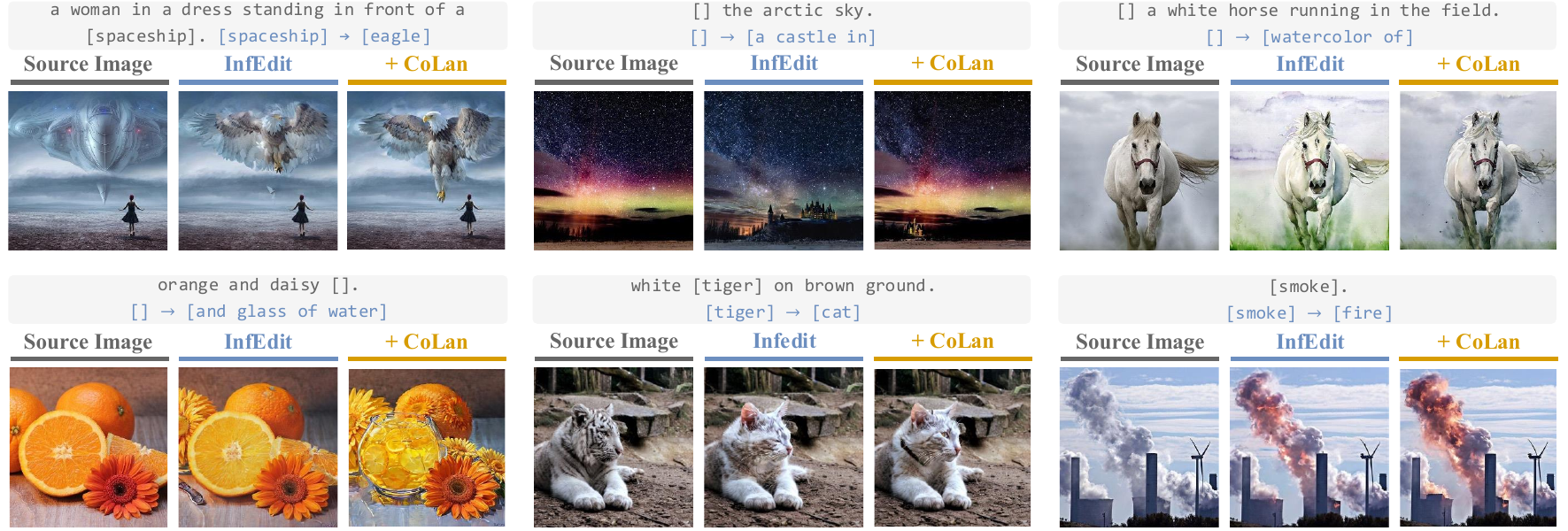}
    \vspace{-6mm}
    \caption{Visual comparisons of \ourframework in the score space (first row) and text embedding space (second row) of InfEdit.  \texttt{\textcolor{jinqidarkgray}{Texts in gray}} are the original captions of the source images from PIE-Bench, and \texttt{\textcolor{jinqidarkblue}{texts in blue}} are the corresponding edit task (replace, add, remove).}
    \label{fig:visual_comparison_infedit}
    \vspace{-1mm}
\end{figure*}

Ideally, the provided editing should be accurately reflected in the region of interest. We observe that editing with the backbone alone often results in either exaggerated or understated editing. For example, in the task of modifying from \texttt{[spaceship]} to \texttt{[eagle]} in Figure~\ref{fig:visual_comparison_infedit} (caption: \texttt{``a woman in a dress standing in front of a [spaceship]''}), the InfEdit backbone alone yields an edited image where the region of interest only resembles an ambiguous bird, whereas an eagle is clearly visible when plugging with \ourframework. Moreover, in Figure~\ref{fig:visual_comparison_p2pzero}, the example with the caption \texttt{``a [meerkat] puppy wrapped in a blue towel.''} has a blue towel wrapped around the meerkat in the source image. With the P2P-Zero backbone alone, the towel is missing from the output image, whereas the output after plugging \ourframework has the blue towel in nearly the same position as that in the source image. 

\begin{table}[t]
    \centering
    \footnotesize
    \caption{\label{tab:time_efficiency} Average time of sparse decomposition in \ourframework for different backbones.}
    \vspace{-2mm}
    \begin{tabular}{ccccc}
        \toprule
         Backbone & \makecell{Metric} & \makecell{Time\\(s)} & \makecell{Proportion\\(\%)} \\

\midrule
        \multirow{2}{*}{P2P-Zero} & Editing Process   & 38.74 & 100   \\
                                  & Sparse Decomposition  &  0.153 & 0.394  \\
        \midrule
        \multirow{2}{*}{Infedit (S)} & Editing Process   & 2.198 & 100   \\
                                  & Sparse Decomposition  & 0.084 & 3.82  \\
        \bottomrule
    \end{tabular}
    \vspace{-2mm}
\end{table}

\begin{figure*}[t]
    \centering
    \includegraphics[width=\linewidth]{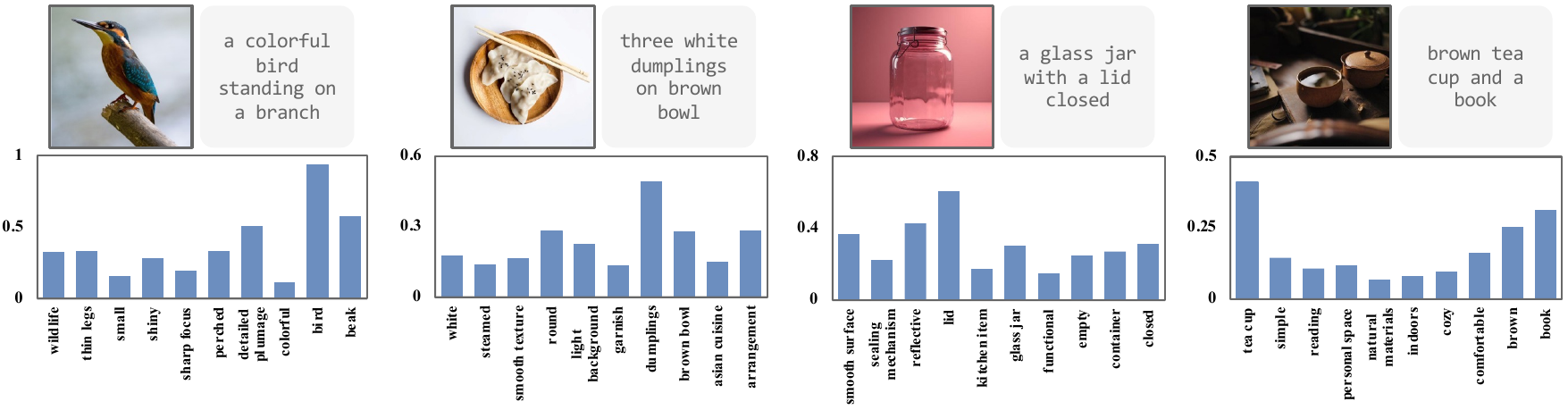}
    \vspace{-6mm}
    \caption{The histograms of solved magnitudes of the concept atoms in \ourframework decomposition (score space). The histogram includes the concepts whose \ourframework coefficients have the top 10 largest magnitudes.}
    \label{fig:coefficient_score}
    \vspace{-1mm}
\end{figure*}

As seen, for both the regions of interest and backgrounds, edited images are of higher quality when a backbone method runs with \ourframework. We postulate that this is possible because \ourframework respects the geometry of the concept vectors via sparse decomposition. By identifying the correct coefficients, our concept transplant is precise and does not significantly affect non-targeted semantics.

\subsubsection{Representation Decomposition}

One of the key steps in our approach (discussed in \S\ref{sec:method-transplant}) is to linearly decompose the latent representation (from the editing backbone) into a sparse combination of dictionary atoms. 
The success of our downstream editing task hinges on finding a proper set of concepts coefficients that accurately reflects the semantics in the source image. Here we verify that \ourframework indeed finds and analyzes representative concepts that are visibly contributive to the given image. 

Figures~\ref{fig:coefficient_clip} and~\ref{fig:coefficient_score} present the magnitude histograms of the concept coefficients solved by \ourframework in the CLIP space and score space respectively. For decompositions in the score space (Figure~\ref{fig:coefficient_score}), take as an example on the leftmost captioned \texttt{``a colorful bird standing on a branch''}. \ourframework finds the top three concepts in the image including ``bird'', ``beak'', and ``detailed plumage'', all of which are concepts relevant to the bird in the provided image. Similarly, take the second image captioned \texttt{``an orange cat sitting on top of a fence''} in Figure~\ref{fig:coefficient_clip}. The top concepts in the histogram are key semantics including ``cat'', ``fence'' and ``orange''. 
Overall, in both spaces, \ourframework is able to find descriptive concepts and solve coefficients to accurately reflect the composition of semantics.

\begin{table}[t]
    \centering
    \footnotesize
    \setlength{\tabcolsep}{3.2pt}
    \caption{\label{tab:dictionary_size} Average performance of backbones with \ourframework for different dictionary sizes ($N$).}
    \vspace{-2mm}
    \begin{tabular}{cccccc}
        \toprule
         Backbone & \makecell{Metric} & \makecell{$N = 5$} & \makecell{$N = 10$} & \makecell{$N = 20$} & \makecell{$N = 30$} \\

\midrule
        \multirow{2}{*}{P2P-Zero} & LPIPS ($\times10^{-3}, \downarrow$)   & 135.6 & 107.1 & 80.12 & 72.85    \\
                                  & Target Concept  & 20.83 & 20.99 & 21.10 & 21.14   \\
        \midrule
        \multirow{2}{*}{Infedit (S)} & LPIPS ($\times10^{-3}, \downarrow$)   & 56.28 & 55.87 & 53.96 & 53.11  \\
                                  & Target Concept  & 22.05 & 22.09 & 22.38 & 22.40  \\
        \bottomrule
    \end{tabular}
    \vspace{-2mm}
\end{table}

\vspace{-1mm}
\subsection{Representation Analysis in \ourdataset}\label{sec:rep-analysis}
\vspace{-1mm}

This section studies the concept vectors obtained from diverse concept stimuli of our \ourdataset dataset. We evaluate the grounding of the concept vectors in \S\ref{sec:concept-grounding} and the variability of the concept in the edited images in \S\ref{sec:coefficient-slider}. 

\subsubsection{Concept Grounding}\label{sec:concept-grounding}

An extracted concept vector is \textit{grounded} when the vector serves effectively in the editing backbone to impose the corresponding visual semantics in the image. For instance, if we use representation reading \cite{parmar2023pix2pixzero,alex2023actadd,zou2023representation,luo2024pace} to convert the stimuli under \texttt{[green]} to the concept vector, then we would expect to see the color `green' as we add this vector in the image editing backbone.

We verify that our concept vectors are grounded in the following way. For a given concept \texttt{[x]}, we extract its concept vector from \ourdataset. Then we generate the edited images by adding the concept vector in the backbone for every source image. Lastly, we evaluate the difference between the CLIP(source image, \texttt{``x''}) and CLIP(edited image, \texttt{``x''}). If the given concept vector is indeed grounded, we would expect to see an increase in the metric. In Table~\ref{tab:concept_grounding}, we sample three of the concept directions \texttt{[watercolor]}, \texttt{[dog]}, \texttt{[wearing hat]}, and apply P2P-Zero with \ourframework to every source image in PIE-Bench. We further divided the results based on the four image types: Artificial, Natural, Indoor, and Outdoor. Across all image types and our given concepts, we observe a significant increase in the CLIP similarity, which means that the edited images are indeed towards the desired concept direction, and the concept vectors are grounded. The results with more concepts and visualization can be found in Appendix \S\ref{sec:appendix_additional_results}.

\begin{figure}[t]
\centering
  \includegraphics[width=\linewidth]{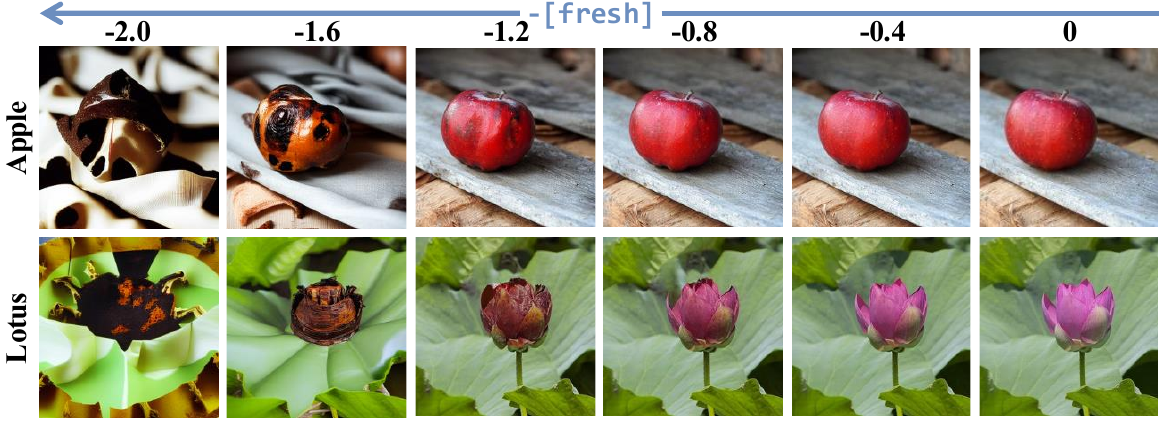}
\caption{Visualizations of edited images with decreasing strength of the concept \texttt{[fresh]} extracted from our \ourdataset dataset. The values on top correspond to the coefficient $\vw_{\text{fresh}}$ for removing the concept $\vd_{\text{fresh}}$. \ourframework solves $\vw^*_{\text{fresh}}$ of $-0.977$ for the apple and $-1.16$ for the lotus.}\label{fig:concept_slider_negative}
\vspace{-1mm}
\end{figure}

\subsubsection{Comparing Editing Strengths}
\label{sec:coefficient-slider}
\begin{figure}[t]
\centering
  \includegraphics[width=\linewidth]{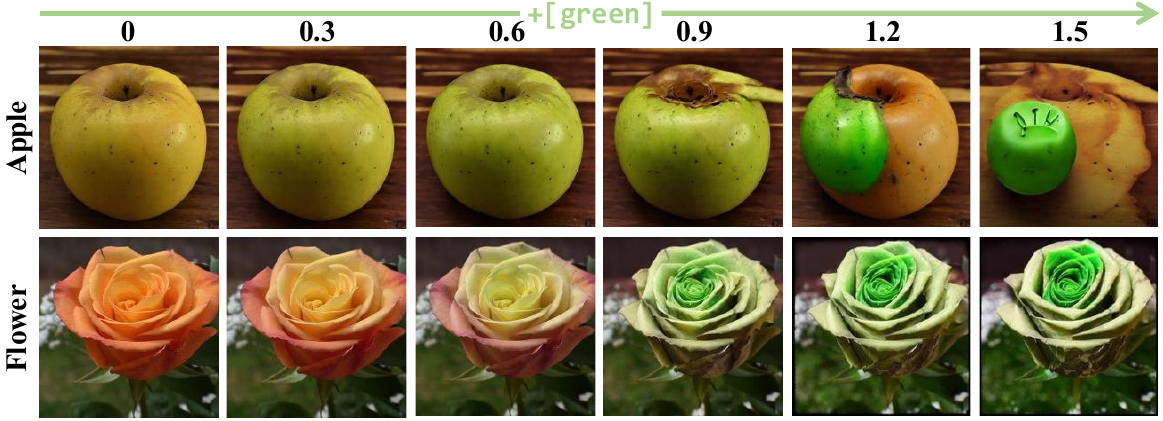}
\caption{Visualizations of edited images with increasing strength of the concept \texttt{[green]} extracted from our \ourdataset dataset. The values on top correspond to the coefficient $\vw_{\text{green}}$ for adding the concept vector $\vd_{\text{green}}$. \ourframework solves $\vw^*_{\text{green}}$ of $0.586$ for the apple and $0.695$ for the rose.
}\label{fig:concept_slider}
\vspace{-1mm}
\end{figure}

As we argued in \S\ref{sec:basics}, proper image editing requires well-estimated edit strength that depends on the presence of concepts in the given source image. Visualizing the progressive changes of the source image along the desired edit direction~\cite{2021StyleCLIP,g2023conceptslider,shen2021closedform,He_2019_attgan} offers insights for estimating edit strength. Here we compare the editing effects of the concept vectors from our \ourdataset dataset with grids of coefficients. Figure~\ref{fig:concept_slider_negative} and Figure~\ref{fig:concept_slider} experiment with two scenarios: concept removal and concept addition, respectively.

Take the top row of Figure~\ref{fig:concept_slider} as an example. Our task here is to add the target concept \texttt{[green]} to our source image of an apple. Our method \ourframework solves the concept coefficient $\vw^*_{\text{green}} = 0.586$. Comparing to edited images by a range of $\vw_{\text{green}}$, we observe that the edited images with $\vw_{\text{green}}  > 0.586$ gradually appear over-edited and corrupted, whereas images with $0 < \vw_{\text{green}} < 0.586$ are still under-edited for the target concept. More concretely, we observe that the green color is not visible at $\vw_{\text{green}} = 0$ and is visible at $\vw_{\text{green}} = 0.6$. Eventually, a brown patch appears on top at $\vw_{\text{green}} = 0.9$, and the apple morphed into a corrupted object at $\vw_{\text{green}} = 1.5$. Similarly, for concept removal in the second row Figure~\ref{fig:concept_slider_negative}, our method \ourframework solves the concept coefficient $\vw^*_{\text{fresh}} = -1.16$. We observe that result images of the lotus with $\vw_{\text{fresh}} < -1.16$ appear over-edited, whereas those with $0 > \vw_{\text{fresh}} > -1.16$ are under-edited. In summary, our results demonstrate that a suitable choice of the strength is important for high-quality image editing, and \ourframework outputs a solution that has edit effectiveness while preserving the visual consistency.

\begin{table}[t]
    \centering
    \footnotesize
    \caption{\label{tab:concept_grounding} Grounding of sampled concept directions in \ourdataset.}
    \begin{tabular}{ccccc}
        \toprule
         \makecell{Concept\\Direction} & \makecell{Image\\Type} & \makecell{Source} & \makecell{Edited} & \makecell{Increase \\(\%, $\uparrow$)} \\

\midrule
        \multirow{4}{*}{\texttt{[watercolor]}} & Artificial   & 15.20 & 18.08 & 18.95  \\
                                  & Natural  &  12.37 & 18.31 & 48.02  \\
                             & Indoor & 12.94 & 16.69 & 28.98  \\
                             & Outdoor & 14.19 & 19.08 & 34.46  \\
        \midrule
        \multirow{4}{*}{\texttt{[dog]}} & Artificial   & 14.18 & 19.28 & 35.97  \\
                                  & Natural  &  13.28  & 18.65 & 40.49  \\
                             & Indoor & 12.46 & 18.29 & 46.81 \\
                             & Outdoor & 13.08 & 18.35 & 40.29 \\
        \midrule
        \multirow{4}{*}{\texttt{\makecell{[wearing\\hat]}}} & Artificial   & 12.58 & 14.77 & 17.41  \\
                                  & Natural  &  11.73 & 14.02 & 19.49  \\
                             & Indoor & 10.25 & 12.18 & 18.83 \\
                             & Outdoor & 11.28 & 13.34 & 18.28 \\
        \bottomrule
    \end{tabular}
    \vspace{-1mm}
\end{table}

\section{Conclusion}
This paper presents Concept Lancet (CoLan), a zero-shot, plug-and-play framework for principled representation manipulation in diffusion-based image editing. By leveraging a large-scale curated dataset of concept representation (\ourdataset), we extract a contextual dictionary for the editing task and perform sparse decomposition in the latent space to accurately estimate the magnitude of concept transplant. Image editing backbones plugging with \ourframework achieve state-of-the-art performance in editing tasks while better maintaining visual consistency. Through extensive quantitative and qualitative evaluations across multiple perspectives, we demonstrate CoLan’s strong capability to interpret and improve the image editing process. We provide further discussions on limitations, future developments, and societal impacts in Appendix \S\ref{sec:appendix_limitation_future_work}.
\clearpage

\section*{Acknowledgment}

This research is supported in part by the Office of the Director of National Intelligence (ODNI), Intelligence Advanced Research Projects Activity (IARPA), via a grant of the BENGAL Program and a grant of the HIATUS Program (Contract \#2022-22072200005), by the Defense Advanced Research Projects Agency's (DARPA) SciFy program (Agreement No. HR00112520300), and by the Penn Engineering Dean's Fellowship. The views expressed are those of the author and do not reflect the official policy or position of the Department of Defense or the U.S. Government.

{
    \small
    \bibliographystyle{ieeenat_fullname}
    \bibliography{main}
}
\clearpage

\clearpage
\setcounter{page}{1}
\maketitlesupplementary

\noindent We organize the supplementary material as follows. 
\begin{itemize}[wide,itemindent=5pt]
    \item \S \ref{sec:appendix_prior_art} covers additional details of prior arts on diffusion-based editing to complement those mentioned in \S \ref{sec:basics}.
    \item \S \ref{sec:appendix_implementation_details} provides descriptions for collecting the dataset \ourdataset and implementing concept transplant method \ourframework.
    \item \S \ref{sec:appendix_additional_results} gives extra visualizations on the dataset and the method. 
    \item \S \ref{sec:appendix_limitation_future_work} discusses limitations, future works and societal impacts.
    \item \S \ref{sec:appendix_prompting_template} details the prompting templates used in dataset collection and inference.
\end{itemize}

\section{Prior Arts for Diffusion-Based Editing}
\label{sec:appendix_prior_art}

To generate a new image $\vz_0$ based on text prompts, diffusion models sample from a standard Gaussian $\vz_T$ and recursively denoise it through the reverse process~\cite{ddim}:
\begin{align}
    \vz_{t-1}=&\sqrt{\alpha_{t-1}}\vf_{\btheta}(\vz_t,t,\vc)+\sqrt{1-\alpha_{t-1}-\sigma_t^2}\bepsilon_{\btheta}(\vz_t,t,\vc) \nonumber \\
    &+ \sigma_t \bepsilon_t,\ \text{ with } \ \bepsilon_t\sim\mathcal{N}(0,\mI)\,.\label{eq_denoise_appendix}
\end{align}
Here $\vz_t$ is the denoised image at time $t$, $\vc$ is the text embedding of the caption of the image to be sampled,
$\bepsilon_{\btheta}(\vz_t,t,\vc)$ and $\vf_{\btheta}(\vz_t,t,\vc)$ are two networks that predict the \emph{score function}~\cite{song2021scorebased} and the denoised image $\vz_0$ respectively given $\vc$ and $\vz_t$. As we elaborate below, different choices for $\alpha_t,\sigma_t,\vf_{\btheta}$ give rise to a class of diffusion models for editing.

\begin{figure*}[t]
    \centering
    \includegraphics[width=\linewidth]{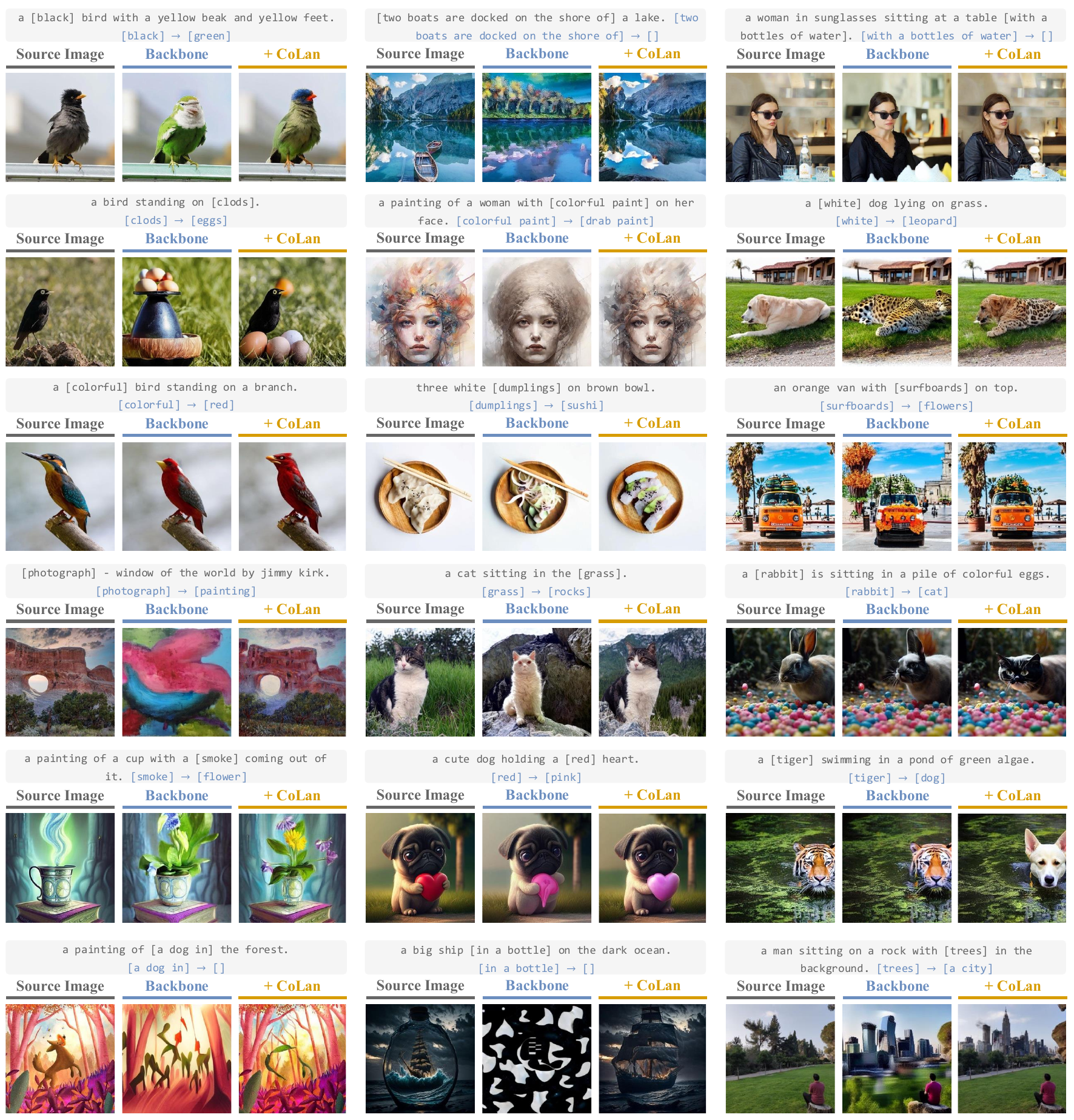}
    \vspace{-5mm}
    \caption{Additional visual comparison of \ourframework in the text embedding space of P2P-Zero. We observe that the backbone plugging with \ourframework has editing results that visually better align with the task.}
    \label{fig:visual_comparison_clip_appendix}
    \vspace{-3mm}
\end{figure*}

\myparagraph{DDIM-Based Editing} By choosing $\sigma_t=0$ and $\vf_{\btheta}(\vz_t,t,\vc)=\frac{\vz_t-\sqrt{1-\alpha_{t}}\bepsilon_{\btheta}(\vz_t,t,\vc)}{\sqrt{\alpha_t}}$ for every $t$, the denoising process in \eqref{eq_denoise_appendix} yields DDIM sampling~\cite{ddim}.
To make sure 
such a process generates the source image $\vz_0$ faithfully, one replaces the stand Gaussain $\vz_T$ with noise computed from a special forward process that iteratively adds deterministic noises, computed via $\bepsilon_{\btheta}(\cdot,t,\vc)$, to the source image $\vz_0$. Some regularization can improve the statistical properties of these noises, resulting in better image editability during the denoising process~\cite{parmar2023pix2pixzero}. Recently, the work of \cite{ju2023directinversion} have proposed \emph{Direct Inversion (DI)} to add further guidance,
allowing exact recovery of $\vz_0$ following the source branch and then improving the visual quality of the edited image when concept transfer is imposed.

\myparagraph{Consistency-Model-Based Editing} Instead of parameterizing $\vf_{\btheta}$ using the learned score $\bepsilon_{\btheta}$, one can learn a separate network for $\vf_{\btheta}(\vz_t,t,\vc)$ to approximate the flow map of the \emph{probablity flow ODE}~\cite{song2021scorebased}, the deterministic counterpart of DDPM~\cite{ddpm} sampling. With the above and the choice of $\sigma_t=\sqrt{1-\alpha_{t-1}}$ for every $t$, the process in \eqref{eq_denoise_appendix} gives \emph{Multi-step Consistency Model Sampling}~\cite{song2023consistency}, and $\vf_{\btheta}(\vz_t,t,\vc)$ in this case is called the \emph{Consistency Model} \cite{song2023consistency,luo2023latentconsistency}. Through a trained consistency model, one can ideally denoise $\vz_t$ into $\vz_0$ in one pass of $\vf_{\btheta}$. However, the denoised $\vz_0^{(t)}:=\vf_{\btheta}(\vz_t,t,\vc)$ has low quality if $\vz_t$ is close to a Gaussian, thus a multi-step sampling 
is adopted to improve the sampled image quality~\cite{song2021scorebased}. For the image editing purpose,~\cite{xu2023inversion} propose \emph{Virtual Inversion (VI)} that guides the process
to sample the source image at every time $t$ in the source branch, i.e., $\vz_0^{(t)}=\vz_0, \forall t$. 

\begin{algorithm*}
    \caption{Concept Lancet (\ourframework) for Diffusion-based Image Editing}
    \DontPrintSemicolon
    \KwIn{Frozen diffusion-based image editing backbone $\mathcal{F}_{\theta}$, image editing tuples (source prompt, source image, target prompt)  $P = \{(\vp_i,\vq_i,\vp'_i)\}_{i=1}^{N_q}$}
    Parse $P$ with the vision-language model to collect the concepts $X = \operatorname{VLM}(P)$\\
    For each concept $x_i$ $\in$ $X$ :{\small\textcolor{lightgray}{\Comment*[r]{\S 3.1:  Concept Dictionary Synthesis}}}
    \Indp
    Instruct the LLM to synthesize concept stimuli $\{\vvs^{x_i}_j\}_{j=1}^{K} = \operatorname{LLM}(x_i)$ \\
    Extract the concept vector $\vd_{x_i} = \operatorname{RepRead}((\vvs^{x_i}_j\}_{j=1}^{K})$ \\
    \Indm
    Stack concept vectors $\{\vd_{x_i}\}_{i=1}^{N_x}$ as columns of the concept dictionary $\mD$.\\
    For each source prompt-image pair $(\vp_i,\vq_i) \in P$:  {\small\textcolor{lightgray}{\Comment*[r]{\S3.2 Concept Transplant via Sparse Decomposition}}}
    \Indp
        Encode $\vp_i$ to the text embedding space or $(\vp_i,\vq_i)$ to the diffusion score space as the source representation $\vv$\\
        Solve for the compositional coefficients that reconstruct the source
        $\vw^* = \argmin_{\vw} \left\|\vv - \mD\vw\right\|_2^2 + \lambda\left\|\vw\right\|_1$\\
        Curate a modified dictionary $\mD'$ by replacing the column of the source concept with that of the target concept\\
        Obtain the edited latent representation as $\vv' = \mD'\vw^* + \vr$.\\
        Generate the edited image through the image editing backbone $q'_i = \mathcal{F}_{\theta}(\vv')$.\\
    \Indm
    \KwOut{The edited images $Q'=\{\vq'_i\}_{i=1}^{N_q}$.}
    \label{alg:colan}

\end{algorithm*}

\begin{figure*}[t]
\vspace{-4mm}
\centering
  \includegraphics[width=\linewidth]{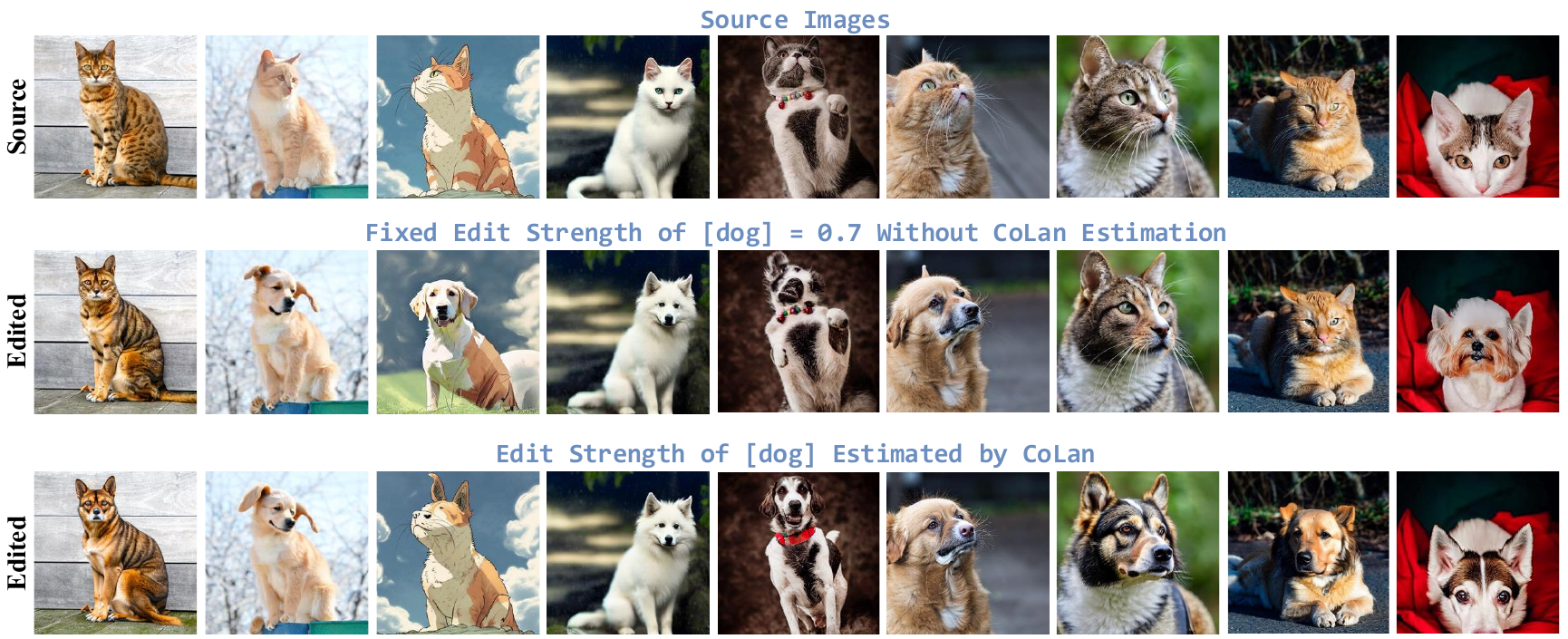}
\caption{Visualizations of editing results. The first row shows the source images, the second row shows the results with the fixed edit strength of $0.7$ for the concept \texttt{[dog]} without \ourframework analysis, and the third row shows the edit results with \ourframework analysis.}
\label{fig:concept_coeff_fix}
\vspace{-4mm}
\end{figure*}

\section{Framework Details}
\label{sec:appendix_implementation_details}

\myparagraph{Dataset Collection} Each concept in the \ourdataset approximately consists of $30$ stimuli. We use GPT-4o (with vision module) \cite{openai2023gpt4} for parsing source input and proposing the concepts. After curating all concepts, we use GPT-4o (without vision module) to generate diverse concept stimuli. The instructions for them are shown in~\S\ref{sec:appendix_prompting_template}.

\myparagraph{Concept Transplant}
When constructing the dictionary in the CLIP text embedding space, each concept vector is a sequence of tokens flattened as a single vector of dimension $d = 77 \times 768 = 59136$, where $77$ is the maximum number of tokens after padding and $768$ is the dimension of token embeddings. For plugging \ourframework on the text embedding space of P2P-Zero, we refer to analyzing the process of $c + \Delta c$ in Algorithm 1 of \cite{parmar2023pix2pixzero}. For plugging \ourframework on the text embedding space of InfEdit, we refer to decomposing the embedding of its source branch to solve the coefficients. For plugging \ourframework on the score space of InfEdit, we refer to analyzing the $\varepsilon_{\tau_n}^{\mathrm{cons}} + \varepsilon_{\tau_n}^{\mathrm{tgt}}-\varepsilon_{\tau_n}^{\mathrm{src}}$ in Algorithm 2 of \cite{xu2023inversion}. Specifically, given a concept $x$, its direction $\vd_x$ for concept dictionary in the score space at the time step $t$ is generated as follows:
$$
\begin{aligned}
\epsilon_x &= \bepsilon_{\btheta}(\cdot;t,\operatorname{RepRead}({E(\vvs_1^x), \dots, E(\vvs_K^x)}))
\end{aligned}
$$
where the $\operatorname{RepRead}(\cdot)$ corresponds to the representation reading algorithms described in \S\ref{sec:method-gendict}.

\myparagraph{Evaluation Detail} In Table~\ref{tab:concept_lancet_performance}, we evaluate all diffusion-based editing baselines with the backbone of Stable Diffusion V1.5 \cite{rombach2021highresolution}, and consistency-based baselines with the Latent Consistency Model \cite{luo2023latentconsistency} (Dreamshaper V7) which is distilled from Stable Diffusion V1.5. The hyperparameter for the sparsity regularizer $\lambda = 0.01$. The null embedding or $\emptyset$ in the paper refers to the CLIP embedding of the empty string. When adding/inserting a target concept, as there is no counterpart described in the source caption, we instruct the VLM to propose a counterpart present in the source image and revise the source caption. The revised dataset will be open-sourced together with all concept stimuli. We use P2P-Zero as the backbone for the representation analysis in \ourdataset and comparing editing strengths. The experiments in \S\ref{sec:exp} are performed on a workstation of 8 NVIDIA A40 GPUs.

\myparagraph{Pipeline} Algorithm~\ref{alg:colan} shows the full procedure of our proposed framework \ourframework. The first part of the algorithm is to extract a set of concept vectors from the input editing image-text tuples based on \S~\ref{sec:method-gendict}), followed by the second part where we transplant the target concept via sparse decomposition in \S~\ref{sec:method-transplant}. In the first part, we instruct a VLM to parse the source input into a set of relevant concepts, and then we instruct an LLM to generate concept stimuli for every concept. Using the concept stimuli, we extract a collection of concept vectors using representation reading from the latent space of our diffusion model. Then, in the second part of \ourframework, we decompose the text embedding or diffusion score of the source representation using sparse coding techniques. After obtaining the coefficients of each concept vector, we perform a transplant process with the customized operation of removing, adding, or replacing. Finally, we synthesize the edited images with the modified latent representation with the image editing backbone.

\section{Additional Results}
\label{sec:appendix_additional_results}

This section provides additional results for \ourframework. It includes more editing improvements with baseline models and visualization of concept instances from our \ourdataset dataset.

\myparagraph{Visual Comparison}
Figure~\ref{fig:visual_comparison_clip_appendix} shows additional visualization of the image editing results. The experiment settings follow \S\ref{sec:qualitative}. We observe that the editing backbone has a better editing performance after plugging \ourframework.

\begin{figure*}[t]
\centering
  \includegraphics[width=\linewidth]{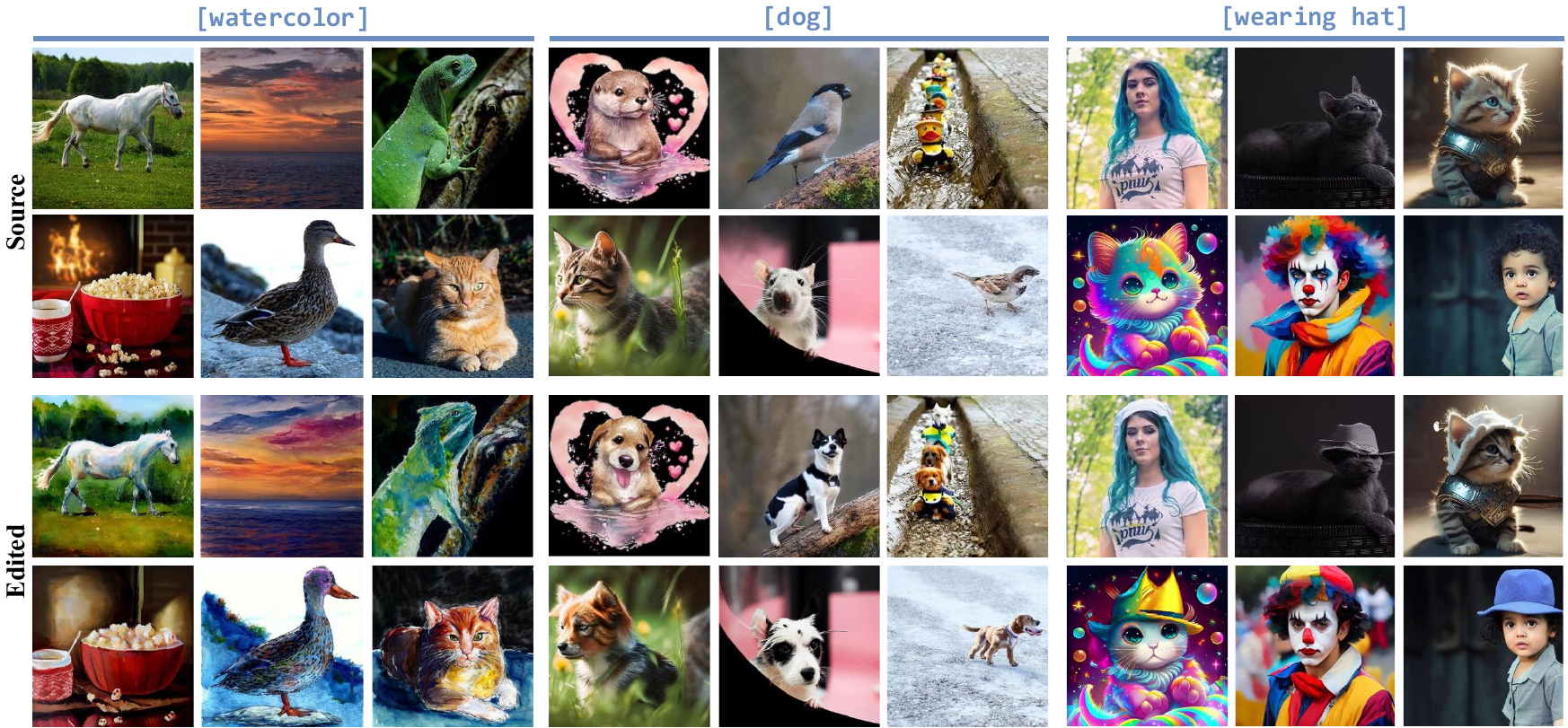}
\caption{Visualizations of concept grounding for sampled concepts from our \ourdataset dataset. We observe that the extracted concept vectors from our dataset corresponds to the desired semantics by visualization.}
\label{fig:concept_grounding}
\vspace{-5mm}
\end{figure*}

\begin{figure*}[t]
    \centering
    \includegraphics[width=\linewidth]{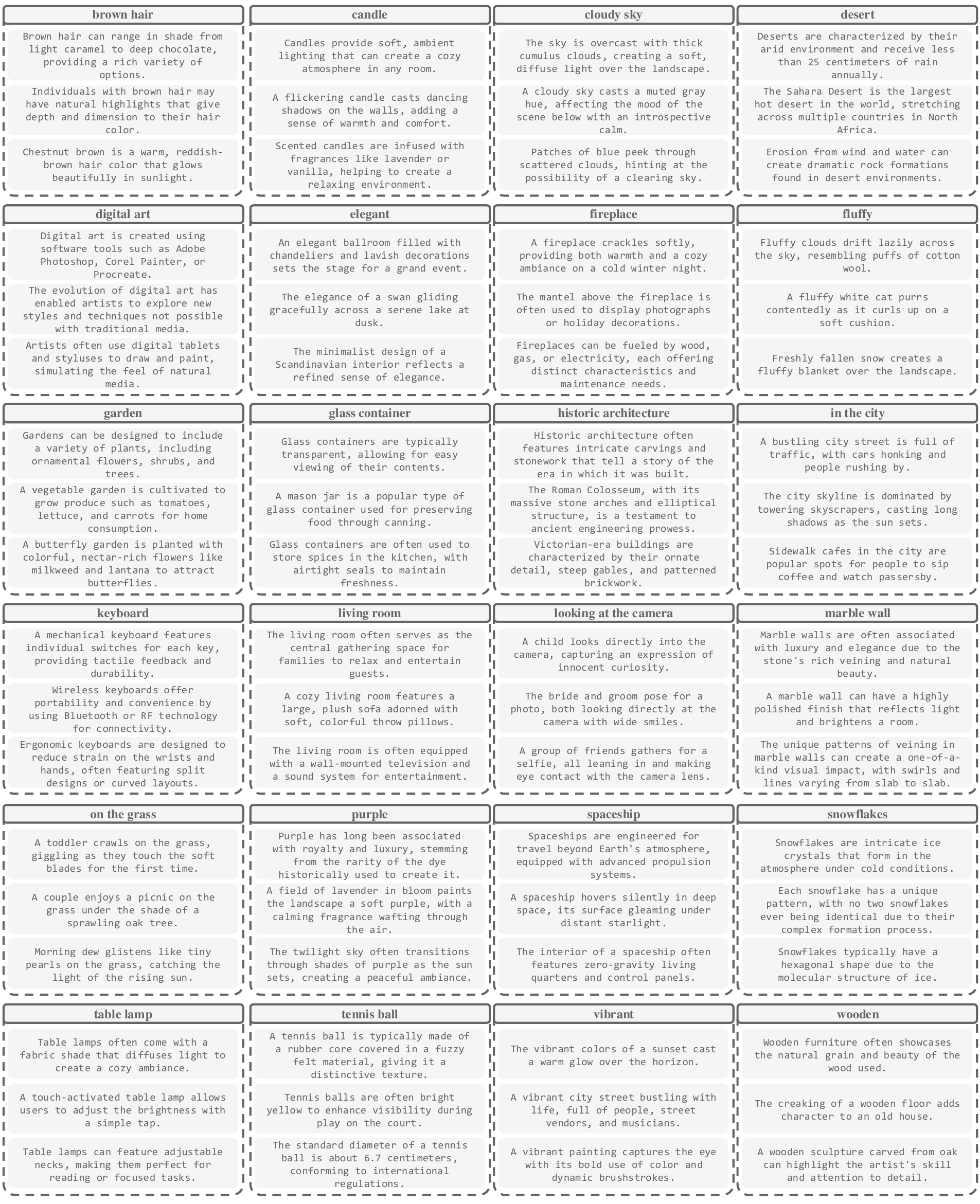}
    \vspace{-6mm}
    \caption{Additional samples of the concept stimuli from \ourdataset. Each concept consists of approximately $30$ stimuli and this figure samples the first three for a concept.}
    \label{fig:concept_stimulus_appendix}
    \vspace{-4mm}
\end{figure*}

\myparagraph{Concept Grounding}
Figure~\ref{fig:concept_grounding} visualizes the edited images with the extracted concept vectors \texttt{[watercolor]}, \texttt{[dog]}, and \texttt{[wearing hat]} from the stimuli of our \ourdataset dataset. We observe that the edited images correctly reflect the semantic meaning of the concepts, which indicates that our concept stimuli successfully ground the concept. Figure~\ref{fig:concept_stimulus_appendix} further shows additional samples of concepts and their stimuli.  Note that there are approximately $30$ stimuli per concept, and our figure shows the first three for each concept.

\myparagraph{Edit Strength} Figure~\ref{fig:concept_coeff_fix} shows the editing results from source images of the cat to the target concept dog without or with \ourframework. The synthesis setting follows the Comparing Editing Strengths section in \S\ref{sec:exp} and we fix the edit strength to $0.7$ if \ourframework is not used. From the second row of the figure, we observe that different source images of the cat require different magnitudes of editing, and simply choosing a unified strength for all source images will frequently result in unsatisfactory results for different images (under-edit or over-edit). Then in the third row of the figure, we show that editing with \ourframework results in more consistent and reasonable visual results. This is because our framework adaptively estimates the concept composition of each image and solves customized edit strengths for each source image.

\section{Limitations, Future Works, Societal Impacts}
\label{sec:appendix_limitation_future_work}

While \ourframework demonstrates strong performance for diffusion-based image editing, we elaborate on potential limitations and directions for future work in this section.

\myparagraph{Limitation} The current framework primarily operates upon diffusion-based backbones with attention-control mechanisms where source concepts correspond to certain regions of interest. It will be challenging to perform spatial manipulations that require editing across different sectors of attention maps. For instance, consider tasks such as moving the cat from right to left or relocating the table to the corner, which shall require non-trivial operations in the attention modules. Another challenge lies in handling numerical modifications, such as changing the number of objects (e.g., changing an image of two cats to have three cats) or composing numerical relations with multiple different objects (e.g., adding two apples to an image of three bananas).

\myparagraph{Future Work} Future work could explore methods to enhance \ourframework's capabilities to handle spatial relationships and global layout modifications while preserving its precise concept manipulation advantages. For numerical editing, it is worthy exploring the bag-of-words effect of CLIP or how the diffusion model shall encode numerical concepts in a way that straightforward manipulation is permitted \cite{yuksekgonul2022and,wu2024conceptmix,wang2023luciddreaming}. The precise mapping between numerical concepts and their representations in the latent space warrants further investigation to enable more sophisticated counting-based edits. 

\myparagraph{Societal Impact} Image editing frameworks with high user-accessibility (through the prompt-based interface) raise considerations about potential misuse. The ability to perform precise conceptual edits could be exploited to create misleading, controversial, or deceptive content. While our framework focuses on enhancing editing quality, future development should incorporate safeguarding against malicious requests and protecting copyrights in content creation.

\section{Prompting Template}
\label{sec:appendix_prompting_template}

As mentioned in Section~\ref{sec:method-gendict}, we instruct the VLM to perform two tasks: rewriting prompts for concept addition or insertion and constructing detailed concept dictionaries. We then instruct the LLM to synthesize concept stimuli. Figure~\ref{fig:instruction_rewrite_caption_for_concept_addition} shows the instructions (prompting template) for rewriting captions by identifying source concepts and generating re-written prompts tailored for image editing tasks. Figure~\ref{fig:instruction_concept_dictionary} shows the instructions for constructing a comprehensive concept list by parsing multimodal information from the source input. This ensures that the list captures diverse and unique aspects of the source image and prompt. Finally, Figure~\ref{fig:instruction_concept_stimulus} shows the instructions for generating diverse and contextually rich concept stimuli, which enables the mapping to conceptual representations.

\newpage
\begin{figure*}
    \centering
    \footnotesize
    \begin{tcolorbox}[colback=promptlightblue, colframe=darkblue, rounded corners, title=Rewriting Captions for Concept Addition/Insertion, fonttitle=\bfseries]
    \texttt{You are one of the best experts in Generative Models and Concept Learning in the world. You are very good at designing concept dictionary to research the representation in latent space from CLIP or Score-based Generative Models, which have wide applications in image editing. You are a great expert in understanding and parsing multimodal information from a given image. Now, given a source prompt, a target prompt, and a source image, your task is to rewrite the source prompt for the image editing task. Usually, there is a focused pair of concepts in the source prompt and the target prompt to be edited (e.g., "cat" to "dog"). The source concept is usually annotated in the brackets ("[]") in the source prompt. However, in some editing tasks, there is no clear source concept mentioned in the source prompt. Hence, for these tasks, you are required to comprehend the source image and identify the corresponding source concept. After comprehending the source image, you need to generate a re-written source prompt with a clearly annotated source concept. \\
}\\

    \texttt{Here are two demonstrations:}\\
    
    \texttt{Source Prompt: a slanted mountain bicycle on the road in front of a building\\
Target Prompt: a slanted [rusty] mountain bicycle on the road in front of a building\\
Source Concept: ""\\
Target Concept: "rusty"\\
Source Image: (IMG)\\
Re-written Source Prompt: a slanted [new] mountain bicycle on the road in front of a building\\
}\\
\texttt{\space}\\
\texttt{Source Prompt: two birds sitting on a branch\\
Target Prompt: two [origami] birds sitting on a branch\\
Source Concept: ""\\
Target Concept: "origami"\\
Source Image: (IMG)\\
Re-written Source Prompt: two [real] birds sitting on a branch\\
}\\

\texttt{The identified source concept should not be the same as the target concept. The response MUST be with brackets ("[]") around the source concept. You should not use "without" frequently. Try your best to comprehend the image. \\
You should only output the re-written source prompt. DO NOT print anything else such as "Here are ...", "Sure, ...", "Certainly, ...".\\
DO NOT print quotation marks unless necessary. Just return the string.}\\

\texttt{Source Prompt: <input>\\
Target Prompt: <input>\\
Source Concept: <input>\\
Target Concept: <input>\\
Source Image: <input>\\
Re-written Source Prompt: <fill the response here>
}
    \end{tcolorbox}

    \caption{The instructions for rewriting the task of concept addition/insertion with the VLM-found source concept as the counter-part.}
    \label{fig:instruction_rewrite_caption_for_concept_addition}
\end{figure*}

\newpage
\begin{figure*}
    \centering
    \footnotesize
    \begin{tcolorbox}[colback=promptlightblue, colframe=darkblue, rounded corners, title=Concept Dictionary Construction, fonttitle=\bfseries]
    \texttt{You are one of the best experts in Generative Models and Concept Learning in the world. You are very good at designing concept dictionary to research the representation in latent space from CLIP or Score-based Generative Models, which have wide applications in image editing. You are a great expert in understanding and parsing multimodal information from a given image.
Now, given a source prompt, a target prompt, and a source image, your task is to parse the given information into a concept list. The concept list consists of concepts, attributes, objects, and items that comprehensively describe the source image and cover the source prompt. 
Your concept list must have at least 15 concepts. As the concept list is for the task of image editing, there is a focused pair of concepts in the source prompt and the target prompt to be edited. The source concept is usually annotated in the bracket ("[]") in the source prompt. You must put the focused concept in the source prompt as the FIRST atom in the concept list. You must NOT put the focused concept in the target prompt in the concept list.
}\\
    
    \texttt{Here are three demonstrations:}\\
    
    \texttt{Source Prompt: a [round] cake with orange frosting on a wooden plate\\
Target Prompt: a [square] cake with orange frosting on a wooden plate\\
Source Concept: "round"\\
Target Concept: "square"\\
Source Image: (IMG)\\
Concept List: ["round", "cake", "orange", "frosting", "wooden", "plate", "swirl", "creamy", "crumbly", "smooth", "rustic", "natural", "muted", "handmade", "warm", "minimalist", "unfrosted", "botanical", "bark", "inviting", "cozy", "textured", "simple", "organic", "earthy", "soft", "classic", "contrasting", "neutral", "clean"]\\
}\\
\texttt{\space}\\
\texttt{Source Prompt: a painting of [a dog in] the forest\\
Target Prompt: a painting of the forest\\
Source Concept: "a dog in"\\
Target Concept: ""\\
Source Image: (IMG)\\
Concept List: ["a dog in", "painting", "forest", "trees", "leaves", "sunlight", "vibrant colors", "orange hues", "pink trees", "purple plants", "playful", "cartoonish", "nature", "animals", "butterflies", "fantasy", "surreal", "whimsical", "tall trees", "shadows", "depth", "light beams", "foliage", "dynamic", "warm tones", "imaginative", "dreamlike", "motion", "soft textures", "layered composition", "bright atmosphere"]\\
}\\
\texttt{\space}\\
\texttt{Source Prompt: blue light, a black and white [cat] is playing with a flower\\
Target Prompt: blue light, a black and white [dog] is playing with a flower\\
Source Concept: "cat"\\
Target Concept: "dog"\\
Source Image: (IMG)\\
Concept List: ["cat", "black", "white", "blue light", "flower", "playing", "paws", "stone path", "curious", "whiskers", "small", "fluffy", "outdoor", "pink petals", "focused", "nature", "detailed fur", "green stem", "bright", "youthful", "movement", "natural light", "close-up", "gentle", "exploration", "soft shadows", "grass between stones", "alert", "innocent", "delicate"]\\
}

\texttt{The concepts in the list should not be redundant or repetitive. Each concept in the list represents a unique perspective of objects, styles, and contexts. The response MUST be in Python list format.\\
You should have at least 15 concepts in the list. You should only output the Python list. \\
DO NOT print anything else such as "Here are ...", "Sure, ...", "Certainly, ...". Just return the list ["", "", ..., ..., ""].}\\

\texttt{Source Prompt: <input>\\
Target Prompt: <input>\\
Source Concept: <input>\\
Target Concept: <input>\\
Source Image: <input>\\
Concept List: <fill the response here>
}
    \end{tcolorbox}

    \caption{The instructions for the VLM to parse the source image-prompt tuple into the concept list for the concept dictionary.}
    \label{fig:instruction_concept_dictionary}
\end{figure*}

\newpage
\begin{figure*}
    \centering
    \footnotesize
    \begin{tcolorbox}[colback=promptlightblue, colframe=darkblue, rounded corners, title=Concept Stimulus Synthesis, fonttitle=\bfseries]
    \texttt{You are one of the best experts in Generative Models and Concept Learning in the world. You are very good at generating concept stimuli to research the representation in latent space from CLIP or Score-based Generative Models, which have wide applications in image editing. You are a great expert in providing relevant information and scenarios based on a given concept.
Now, given a concept, your task is to generate 30 (THIRTY) instances of concept stimuli for a given concept. As the concept stimuli will be used for the task of image editing, we need comprehensive, diverse, and accurate descriptions and examples for the concept. 
}\\
    
    \texttt{Here are three demonstrations of the concept and its corresponding concept stimuli:}\\
    
    \texttt{Concept: dog}\\
    \texttt{Concept Stimuli:}\\
    \texttt{[\\
    "Dogs are known for their loyalty and strong bonds with humans.",\\
    "A dog wags its tail excitedly when it sees its owner after a long day.",\\
    "Puppies often chew on objects as a way to explore their environment.",\\
    "The sound of a dog’s bark can vary depending on its breed and mood.",\\
    "Dogs rely heavily on their sense of smell, which is far more sensitive than that of humans.",\\
    "A dog runs alongside its owner during a morning jog, full of energy.",\\
    ...\\
]}\\
\texttt{\space}\\
\texttt{Concept: cat}\\
\texttt{Concept Stimuli:}\\
\texttt{[\\
    "Cats are known for their graceful, stealthy movements.",\\
    "A cat stretches lazily under the warm afternoon sun.",\\
    "Kittens explore their surroundings with curiosity and playfulness.",\\
    "A cat’s purring has been shown to have a calming effect on humans.",\\
    "Stray cats often rely on their instincts and sharp senses for survival.",\\
    "The eyes of a cat reflect light in the dark, giving them superior night vision.",\\
    ...\\
]}\\
\texttt{\space}\\
\texttt{Concept: cake}\\
\texttt{Concept Stimuli:}\\
\texttt{[\\
    "Cakes are often baked in layers and filled with frosting or cream in between each layer.", \\
    "A slice of cake reveals its moist interior, topped with a rich layer of chocolate ganache.", \\
    "Cakes are a common centerpiece for celebrations such as birthdays, weddings, and anniversaries.", \\
    "A cake adorned with fresh berries and whipped cream makes for a light, summery dessert.", \\
    "Cupcakes are miniature cakes baked in individual paper liners and often topped with buttercream frosting.", \\
    "The aroma of a freshly baked vanilla cake fills the kitchen with a warm, sweet scent.", \\
    ...\\
]}\\

\texttt{The concept stimuli in the list should not be redundant or repetitive. Each stimulus in the list represents a unique perspective (e.g., styles, contexts, examples, attributes, descriptions, usages) of the concept. The response MUST be in Python list format.\\
You should have at least 30 stimuli in the list. You should only output the Python list. \\
DO NOT print anything else such as "Here are ...", "Sure, ...", "Certainly, ...". Just return the list ["", "", ..., ..., ""].}\\

\texttt{Concept: <input>\\Concept Stimuli: <fill the response here>}
    \end{tcolorbox}

    \caption{The instructions for the LLM to generat diverse stimuli given a concept.}
    \label{fig:instruction_concept_stimulus}
\end{figure*}


\end{document}